%% file: main_arxiv.tex
\documentclass[runningheads]{llncs}

\usepackage[mobile]{eccv}

\usepackage{eccvabbrv}

\usepackage{graphicx}
\usepackage{booktabs}
\usepackage{xspace}
\usepackage{booktabs}
\usepackage{multirow}
\usepackage{multicol}
\usepackage{graphics}
\usepackage{xcolor} 
\usepackage{pifont}
\usepackage{hhline}
\usepackage{makecell}
\usepackage{mdframed}
\usepackage{tcolorbox}
\usepackage{listings}
\usepackage{cuted}
\usepackage{tcolorbox}

\usepackage{enumitem}
\usepackage{colortbl}
\definecolor{cvprblue}{rgb}{0.21,0.49,0.74}
\usepackage{graphicx}
\definecolor{lightcyan}{rgb}{0.88,0.95,1}

\newcommand{\vsp}{\vspace{-2mm}}
\newcommand{\nbf}{\vsp\paragraph}

\usepackage[accsupp]{axessibility}  

\newcommand{\benchmarkname}{HalDec-Bench\xspace}

\usepackage[pagebackref,breaklinks,colorlinks,citecolor=eccvblue]{hyperref}

\usepackage{orcidlink}

\begin{document}

\title{HalDec-Bench: Benchmarking Hallucination Detector in Image Captioning} 

\titlerunning{HalDec-Bench}




\author{
Kuniaki Saito\textsuperscript{1*},
Risa Shinoda\textsuperscript{2*}, 
Shohei Tanaka\textsuperscript{1},
Tosho Hirasawa\textsuperscript{1},\\
Fumio Okura\textsuperscript{3},
Yoshitaka Ushiku\textsuperscript{1} \\
}

\institute{\textsuperscript{1}OMRON SINICX, \textsuperscript{2}The University of Tokyo, \textsuperscript{3}The University of Osaka}
\begingroup
\renewcommand\thefootnote{*}
\footnotetext{Equal contribution. Kuniaki serves as the project lead, while Risa is responsible for dataset construction. Contact: \texttt{kuniaki.saito@sinicx.com}, \texttt{rsihnoda@iis.u-tokyo.ac.jp}.}
\endgroup

\maketitle
\vspace{-20pt}
\input{figures/theaser}
\begin{abstract}
Hallucination detection in captions (\textbf{HalDec}) assesses a vision-language model's ability to correctly align image content with text by identifying errors in captions that misrepresent the image. Beyond evaluation, effective hallucination detection is also essential for curating high-quality image-caption pairs used to train VLMs. However, the generalizability of VLMs as hallucination detectors across different captioning models and hallucination types remains unclear due to the lack of a comprehensive benchmark.
In this work, we introduce \benchmarkname, a benchmark designed to evaluate hallucination detectors in a principled and interpretable manner. \benchmarkname contains captions generated by diverse VLMs together with human annotations indicating the presence of hallucinations, detailed hallucination-type categories, and segment-level labels. The benchmark provides tasks with a wide range of difficulty levels and reveals performance differences across models that are not visible in existing multimodal reasoning or alignment benchmarks. Our analysis further uncovers two key findings. First, detectors tend to recognize sentences appearing at the beginning of a response as \textit{correct}, regardless of their actual correctness. Second, our experiments suggest that dataset noise can be substantially reduced by using strong VLMs as filters while employing recent VLMs as caption generators. 
Our project page is available at \url{https://dahlian00.github.io/HalDec-Bench-Page/}.
\end{abstract}

\input{chapter/1_intro}
\input{chapter/2_related}

\input{chapter/3_dataset}
\input{chapter/4_experiments}

\input{chapter/5_conclusion}

\section*{Acknowledgements}
This work was supported by JST PRESTO, Japan, Grant Number JPMJPR2523. This work was partly achieved through the use of SQUID at D3 Center, The University of Osaka. Also, this work used computational resources TSUBAME4.0 supercomputer provided by Institute of Science Tokyo through Joint Usage/Research Center for Interdisciplinary Large-scale Information Infrastructures and High Performance Computing Infrastructure in Japan.

%
%
\bibliographystyle{splncs04}
\bibliography{main}
\clearpage
\appendix
\input{supplemental_arxiv}

\end{document}

%% file: figures/theaser.tex
\begin{figure*}[h!]
\centering
\includegraphics[width=\linewidth]{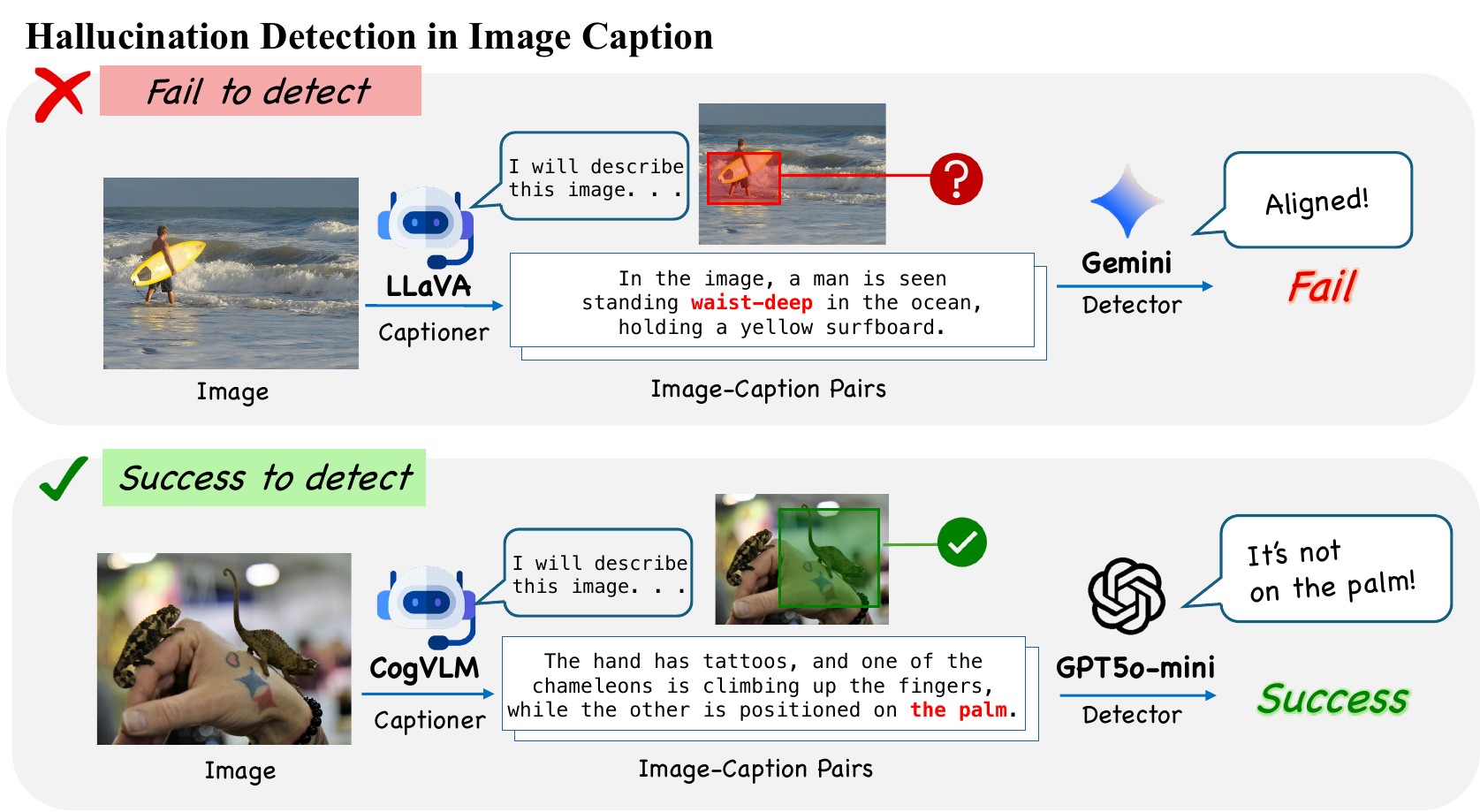}
\vspace{-0.2in}
\captionof{figure}{We introduce a novel benchmark, \textbf{\benchmarkname}, which evaluates the VLM's ability to detect hallucinations in captions. We employ state-of-the-art VLMs to generate image-caption pairs (Captioner) and manually annotate the hallucinated parts when present. We benchmark diverse VLMs as Detector and find that subtle hallucinations can be hard to detect, even by state-of-the-art VLMs.}
\label{fig:teaser}
\vspace{-20pt}
\end{figure*}

%% file: chapter/1_intro.tex
\section{Introduction}

We have seen remarkable progress in large vision-language models (VLMs)~\cite{wang2024qwen2,liu2023visual,liu2024improved,chen2023minigpt,li2023blip}.
A key to this progress lies in understanding image content in the form of text, \ie, learning image-text alignment. Once this mapping between images and text is effectively learned, large language models (LLMs) can be used for various visual reasoning tasks~\cite{li2023blip}.

Hallucination detection in captions, called \textbf{HalDec} hereafter, is a task that assesses VLM's image-text alignment capability. It aims to identify errors in captions that misrepresent image content, such as misstated object counts, incorrect attributes or relationships, or the introduction of entities absent from the image~\cite{biten2022let,li2023evaluating,rohrbach2018object}. Beyond evaluating the alignment ability of VLMs, HalDec enables filtering out unaligned image-caption pairs~\cite{li2022blip} from the training data. In practice, VLM training often relies on captions synthesized by a \textit{Captioner}~\cite{chen2024sharegpt4v,yang2023alip}\footnote{To avoid confusion, we use the term \textit{Captioner} to denote a VLM used for caption generation, and use \textit{Detector} to denote a VLM used for hallucination detection.} to supplement the limited availability of human-annotated data. 
However, these synthetic captions frequently suffer from hallucinations. Curating high-quality image-caption pairs with strong detectors, therefore, plays a crucial role in building performant VLMs~\cite{chen2024sharegpt4v,yu2024your}. Indeed, models such as CLIP~\cite{radford2021learning} and BLIP~\cite{li2022blip} have already been widely used to curate large-scale training datasets for VLMs~\cite{BetkerImprovingIG,li2023blip}. 

Considering the scalability of detectors, we expect the detector to be universally applicable across diverse image-caption pairs. Thus, evaluating HalDec requires testing models to detect hallucinations across different Captioners, image domains, and hallucination types, since each factor can introduce distinct language styles and error patterns. Yet, the universality of large VLMs as a hallucination detector remains unclear although recent studies focus on training a detector model, rather than revealing diverse VLMs ability as a detector~\cite{gunjal2024detecting,wada2025zina}. 

The challenge of making the comprehensive HalDec benchmark is to build a dataset suited for such evaluation, requiring a significant cost of human annotation, where annotators must carefully check the image-sentence alignment. In fact, existing HalDec datasets suffer from limited model coverage and insufficient scale. Rule-based or artificially constructed hallucinations can be unsuitable for analyzing the diverse hallucinations generated by VLMs~\cite{petryketal2024aloha,yuksekgonul2022and}.
MHalDetect~\cite{gunjal2024detecting} provides annotations for only a single VLM, and MHaluBench~\cite{chen2024unified} covers only a small set of models and samples. 
Also, despite the development of benchmarks for multimodal reasoning~\cite{yue2024mmmu,yue2024mmmu-pro,liu2023visual,lu2023mathvista}, a benchmark to evaluate VLMs' fundamental image-caption alignment is limited to hallucinated sentences generated by a human-designed pipeline~\cite{yuksekgonul2022and,hsieh2023sugarcrepe}.\looseness=-1

In this paper, we introduce \benchmarkname, a benchmark designed to evaluate hallucination detectors for image captions in a principled and interpretable manner. The captions in \benchmarkname are generated by diverse VLMs, resulting in a wide range of linguistic styles, vocabularies, and hallucination types. Human annotators carefully examine each sentence to determine the presence of hallucinations, and we provide detailed hallucination-type categories together with segment-level annotations. Beyond serving as a tool for analyzing detectors, \benchmarkname also functions as a testbed for probing VLMs’ fundamental ability to capture image-caption alignment across diverse sentence and image domains. In our experiments, we focus on sentence-level hallucination detection and assess a variety of VLMs as detectors.

The results highlight several advantages of \benchmarkname as a benchmark for hallucination detection in image captions (HalDec).
First, by incorporating captions generated by diverse VLMs, \benchmarkname provides tasks with a wide range of difficulty levels, making it well-suited for evaluating hallucination detectors. Second, the benchmark reveals performance differences across models that are not clearly observable in existing benchmarks, including image reasoning benchmarks such as MMMU and prior image-text alignment benchmarks.

Our analysis further reveals several empirical findings that have not been previously reported. First, detectors tend to recognize sentences appearing at the beginning of a response as \textit{correct}, regardless of their actual correctness. Second, we empirically confirm that dataset noise can be substantially reduced by using strong VLMs as filters across both strong and weak captioners.

%% file: chapter/2_related.tex
\section{Related Work}

\paragraph{Datasets for hallucination detection in image captioning.} 
Some datasets are introduced for HalDec~\cite{wang2023evaluation,chen2024unified,gunjal2024detecting,wada2025zina} (summarized in Table~\ref{tab:stats_other_dataset_cvpr}) and greatly contributed to the development of HalDec models\footnote{ZINA is concurrent with ours, and the dataset was not available at the time of submission; we compare as best we can.}, but they are limited as a HalDec benchmark, often lacking diverse Captioners or sufficient samples per model. Our benchmark, \benchmarkname, addresses these gaps by (i) covering responses from various models, (ii) balancing data across models. SUGARCREPE~\cite{hsieh2023sugarcrepe} and ARO~\cite{yuksekgonul2022and} probe CLIP’s fine-grained image-text alignment ability, but they employ simple sentences and rely on language model or rule-based perturbations to produce hallucinations. In contrast, \benchmarkname uses VLM-generated captions, which are more challenging as shown in Sec.~\ref{sec:experiments}.

\vsp
\paragraph{Hallucination detection and mitigation in image captioning.} 
Hallucination detection in image captioning has been widely studied~\cite{rohrbach2018object}. 
CHAIR~\cite{rohrbach2018object} was the first metric to evaluate image-caption alignment at the object level using an object detector. However, its effectiveness is constrained by the detector’s coverage and accuracy; thus, it fails in capturing the diverse hallucination types and captioning styles. 
Also, many works attempt to mitigate the hallucinations in image captions~\cite{zhang2024vl,leng2024mitigating,farquhar2024detecting,zhou2023analyzing,zhuang2025vasparse,favero2024multi,woo2024don,suo2025octopus}, especially, mitigating hallucinations in long captions is important as they are prone to contain more hallucinations~\cite{zhou2023analyzing,hirota2025lotus}.
Refining a captioning model based on image-caption alignment score, computed by VLMs, is a promising approach~\cite{deng2024seeing}, and our work closely contributes to this line of work. Recent approaches fine-tune VLMs~\cite{gunjal2024detecting} with human-annotated data, calling LLM to leverage tools like open-vocabulary detectors, OCR~\cite{chen2024unified}, or estimate prediction uncertainty~\cite{farquhar2024detecting}. 
Despite these methodological developments and the use of VLMs as a detector, VLMs' fundamental ability to detect hallucinations in captions is unclear due to the lack of a benchmark.

%% file: chapter/3_dataset.tex
\section{Datasets}

\input{tables/stats_other_dataset}
\input{figures/ann_ex}
\vspace{-2mm}
We aim to collect datasets that cover diverse image-caption pairs equipped with high-quality annotations of semantic alignment. 
This section first explains how we collect image-caption pairs and provide annotations, followed by an analysis of the dataset. We focus on obtaining labels for sentence-level semantic alignment for two reasons: (i) sentence-level labels give a cue to easily find more fine-detailed locations of unaligned descriptions, and (ii) span-level annotation suffers more from the subjectivity of annotation than sentence-level. For deeper analysis, we also provide span-level hallucination presence labels and categorize hallucination types. Due to the limited space, we leave most details in the appendix. 

\subsection{Collecting Image-Caption Pairs}
We aim to assess the ability to judge whether a given image-text pair is semantically aligned. To build such a benchmark, it is essential to cover diverse image domains and caption styles. We therefore construct image-caption pairs using ten image-to-text models, where each \textit{caption} consists of multiple consecutive sentences. Although our primary focus is hallucination detection in captions generated by image-to-text models, \benchmarkname additionally includes images generated by five text-to-image models to assess the robustness of detectors across diverse image-sentence pairs. 

\vsp
\paragraph{Image-to-text models (captioner).} We employ CC12M~\cite{changpinyo2021conceptual} and the validation split of COCO 2017~\cite{lin2014microsoft} as image inputs. To ensure diversity in the test images, we cluster images into 50 clusters and pick 40 images from each cluster, resulting in 2000 images in total. We manually categorize each cluster into 12 image domain categories and confirm their diversity. 
We employ GPT-4o, ShareGPT (S-GPT)~\cite{chen2024sharegpt4v}, LLaVA-NeXT (72B)~\cite{li2024llavanext-strong}, Llama-4 (109B)~\cite{llama4}, Qwen2 (7B)~\cite{wang2024qwen2}, and CogVLM (19B)~\cite{wang2024cogvlm}, GPT-5~\cite{singh2025openai}, and Gemini-3-Pro~\cite{gemini-3pro}, covering diverse architectures, scales, and openness. We additionally include Qwen 2.5~(32B)~\cite{qwen2025qwen25technicalreport} and Gemma-3~(27B)~\cite{gemmateam2025gemma3technicalreport} for self-preference analysis. This diversity enables the collection of captions with differing levels of detail and language style. We apply a chat template (\eg, ``Describe the image in detail.'') to get captions. 

\vsp
\paragraph{Text-to-image models.} We additionally construct text-to-image pairs to further evaluate whether detectors can detect hallucinations arising from image generation. Specifically, we employ Stable Diffusion 3.5 (SD)~\cite{sdlarge}, as well as several image generation models accessible through proprietary systems, including GPT-Image-1 (via GPT-4o-mini), Gemini 2.5 Flash Image~(Gemini-2.5), GPT-Image-1.5~(GPT-Img-1.5), and Gemini 3 Pro Image~(Gemini-3). We design a pipeline to ensure diversity in a text prompt and generate 5,500 images in total. See Appendix for more details.

\subsection{Annotation}
In the main paper, we focus on how to obtain labels for sentence-level semantic alignment and leave the fine-detailed annotation process in the appendix. 
We ask annotators to determine if the image presents all the details described in the text correctly. 
The use of SOTA models as captioners makes the annotation non-trivial because hallucinations produced by such models are often subtle and not immediately apparent at first glance, as shown in Fig.~\ref{fig:ex_ann_cvpr}.

\input{figures/pos_haltype}
\vsp
\paragraph{Annotation protocol.} We are inspired by the labeling scheme of \cite{gunjal2024detecting}, where annotators assign one of three categories: \textit{correct}, \textit{incorrect}, or \textit{unknown}. An example of an annotation screen is illustrated in the supplemental material. A sentence is labeled \textit{correct} if it accurately describes the image, and \textit{incorrect} if it contains a part that does not correctly describe the image. When correctness cannot be determined---for example, if the object is too small to recognize or if the description involves non-visible attributes such as smell or wind---it is labeled \textit{unknown}. The \textit{unknown} category is introduced to exclude unreliable cases from evaluation.

\vsp
\paragraph{Annotation process.} To ensure reliable annotations, we adopt a two-stage pipeline consisting of crowd-sourced annotation followed by a review process. In the first stage, workers annotate the presence of hallucinations for each sentence while their performance is continuously monitored through quality checks. Annotators who do not meet the required criteria are replaced to maintain annotation quality. In the second stage, the annotations are reviewed by qualified reviewers to ensure the reliability of the final labels. This process results in a dataset with sentence-level correctness annotations. 
Our dataset also includes segment-level hallucination presence labels and hallucination category labels, where hallucinations are categorized into eight types: \textit{Attribute}, \textit{Object}, \textit{Number}, \textit{Location}, \textit{Illusion}, \textit{Direction}, \textit{Text}, \textit{Relation}, and \textit{Other}. These categories should reveal the weakness of the current VLMs in understanding the image content.

\subsection{Overview of the Dataset}\label{sec:analysis_dataset}
\vspace{-2mm}

\paragraph{Examples of annotated sentences.} Figure~\ref{fig:ex_ann_cvpr} presents unaligned image-sentence pairs. Captioners’ errors often involve visual details or object relationships rather than clear mistakes, making them hard to detect. Therefore, detectors need a fine-grained understanding of image-text alignment. 

\vsp
\paragraph{Basic stats.} 
\benchmarkname contains approximately 104.3K annotated sentences labeled as \textit{correct} or \textit{incorrect}, including 87.5K correct and 16.8K incorrect sentences. The ratio of correct and incorrect sentences varies across captioning models. Nevertheless, \benchmarkname provides a sufficient number of both correct and incorrect sentences for reliable evaluation (excluding \textit{unknown} cases).
As expected, more advanced models are less prone to hallucinations; for example, GPT-5 hallucinates in only 6.5\% of sentences, whereas ShareGPT hallucinates in 26.6\%. Detailed statistics are provided in the supplementary material. Since most models achieve over 90\% accuracy, a large number of generated outputs is necessary to obtain sufficient incorrect samples, particularly for advanced captioners.
Figure~\ref{fig:dataset_details}~(a) shows the distribution of caption lengths, indicating that our dataset covers a wide range of caption lengths.

\vsp
\paragraph{Hallucination categories.} 
We show the stats of hallucination types in Fig.~\ref{fig:dataset_details}~(b). 
Errors in \textit{attributes} are the leading category for many models, indicating that adjectival descriptions (\eg, color, texture) are prone to hallucination as indicated by Fig.~\ref{fig:ex_ann_cvpr}. \benchmarkname provides a large number of sentences, enabling reliable evaluation per hallucination type. 

\vsp
\paragraph{Comparison to existing HalDec datasets.} Table~\ref{tab:stats_other_dataset_cvpr} compares with existing image-text alignment datasets. \benchmarkname substantially surpasses prior datasets in scale, with over 104K sentences—one to two orders of magnitude larger than existing alignment benchmarks. It contains longer, image-caption pairs generated by diverse models~(Fig.~\ref{fig:dataset_details}~(c)), with multiple sentences per image~(Fig.~\ref{fig:dataset_details}~(d)) and a richer vocabulary, enabling evaluation on diverse and long-tail concepts. 

%% file: tables/stats_other_dataset.tex
\begin{table}[t]
\centering
\fontsize{7.0pt}{7.0pt}\selectfont
\setlength\tabcolsep{3.pt} 
\renewcommand{\arraystretch}{1.0} 
\caption{\small \textbf{Comparison with other datasets.}
\benchmarkname evaluates on a substantially broader vocabulary, using real hallucinations naturally produced by VLMs rather than synthetically constructed ones. \benchmarkname includes image-sentence pairs generated by diverse VLMs and records the source model for each pair, enabling detailed analysis of model-specific hallucination patterns.}
\vspace{-3mm}
\begin{tabular}{lcccccc}
\toprule
Dataset & \# Sentences  & Vocabulary &Words/Sent.&\makecell{Hallucination \\ Source}& \makecell{Real \\ Hallucination}&\makecell{Diverse \\ Models}\\
\midrule
Foil~\cite{petryketal2024aloha}        & 5k   & 4.1k &11.8& Rule-based &&\\
HAT~\cite{petryketal2024aloha}         & 0.4k     & 1.2k &13.6& Rule-based& &\\
ARO~\cite{yuksekgonul2022and} & 2k & 0.6k  &7.6&Rule-based&&\\
SugarCrepe~\cite{hsieh2023sugarcrepe}  & 2k   & 2.2k &11.1& LM &&\\
Winoground~\cite{thrush2022winoground}   & 1.6k     & 0.9k  &9.0& Human &&\\
MHalDetect~\cite{gunjal2024detecting}  & 14k  &  4.4k&\textbf{18.0}&VLM& \checkmark&\\
MHaluBench~\cite{gunjal2024detecting}  &  0.7k  & 1.3k&14.6&VLM& \checkmark&\\
\rowcolor{lightcyan}
\benchmarkname & \textbf{104k}  & \textbf{17.5k} &17.8 &VLM& \checkmark&\checkmark\\
\bottomrule
\end{tabular}
\label{tab:stats_other_dataset_cvpr}
\end{table}

%% file: figures/ann_ex.tex
\begin{figure*}[t!]
\centering
\includegraphics[width=\linewidth]{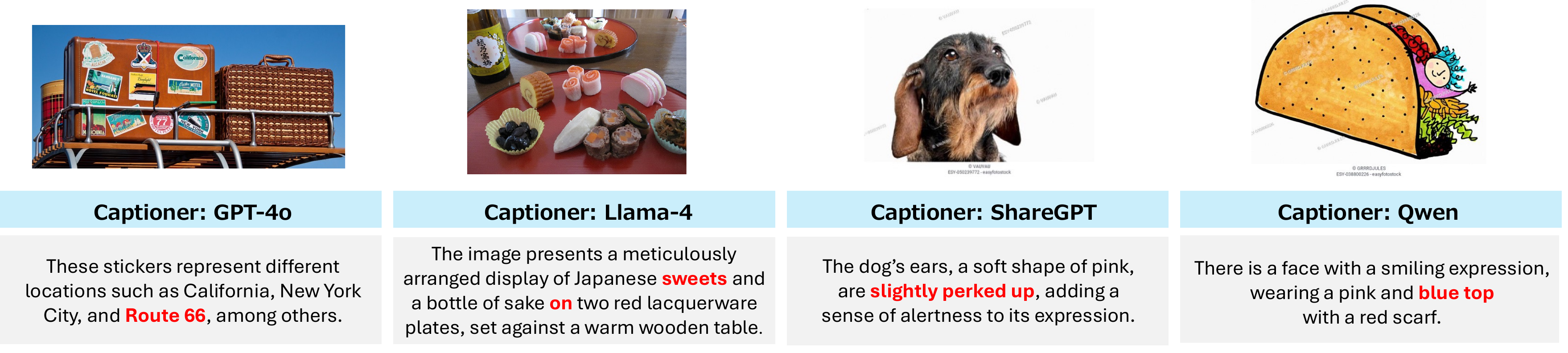}
\vspace{-5mm}
\caption{\small \textbf{Hallucinated sentences in \benchmarkname.} Hallucinated portions are often subtle, requiring fine-grained image-text alignment ability to detect them.}
\label{fig:ex_ann_cvpr}
\end{figure*}

%% file: figures/pos_haltype.tex
\begin{figure*}[t]
\centering
\vspace{-2mm}
\includegraphics[width=\linewidth]{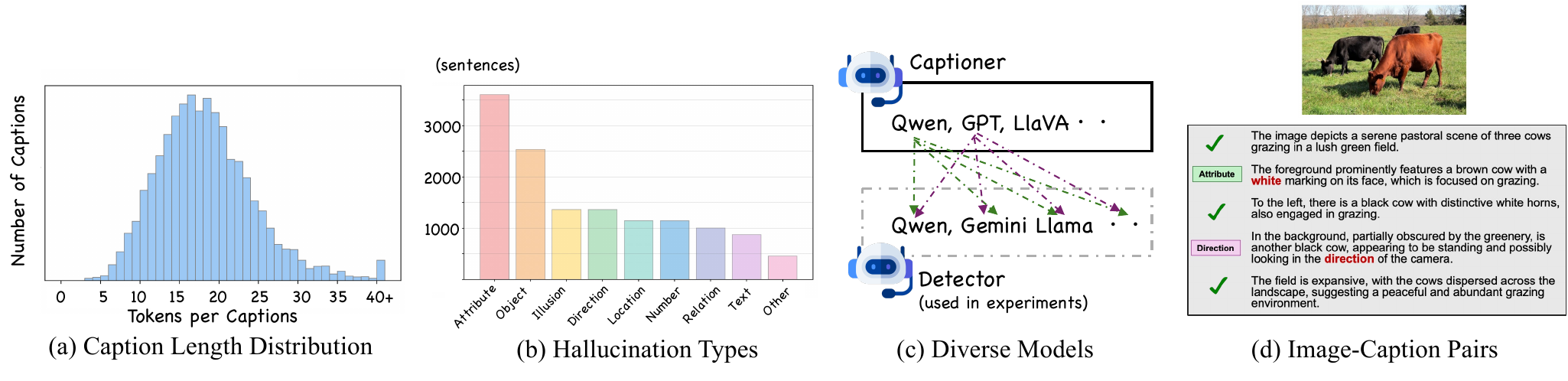}
\vspace{-6mm}
\caption{\small \textbf{Details of our \benchmarkname}. 
(a) We confirm that our dataset includes diverse caption length. 
(b) We also annotate hallucination types into 8 categories. The rarest type is \textit{Text}, but it still includes nearly 500 instances, enabling deeper analysis. 
(c) We create captions using diverse VLMs as Captioners, enabling analysis with combinations of diverse Detector models.
(d) An example of an annotation.}
\label{fig:dataset_details}
\end{figure*}

%% file: chapter/4_experiments.tex
\section{Experiments}
\label{sec:experiments}
\vspace{-2mm}
We aim to benchmark and analyze diverse VLMs on \benchmarkname to uncover key factors for building a performant HalDec model. After describing the experimental setup, we first present an overview of the empirical results, followed by a detailed analysis. 

\vsp
\paragraph{Setups.}
We aim to benchmark diverse VLMs in sentence-level image-text alignment, \ie, the image presents all the details described in the single sentence correctly. Specifically, each sentence and image is independently fed into VLMs. We choose this evaluation protocol since the prior work on hallucination detection~\cite{mishra2024fine} and image-text alignment~\cite{yarom2023you,thrush2022winoground,hsieh2023sugarcrepe} also employs sentence-level evaluation. Following \cite{chan2023clair}, we prompt the decoder-based VLMs to output the score of the alignment between the image and an input sentence, ranging from 0 to 100, as shown in the appendix.
We also include BLIP-2~\cite{li2023blip}, TripletCLIP~\cite{patel2024tripletclip} as fundamental image-text alignment models. Given the alignment score, we compute the AUROC within each captioner, which enables threshold-free evaluation, and the random prediction results in a score of 50. \looseness=-1

Our analysis of \benchmarkname reveals both previously unreported insights and findings that confirm or extend observations suggested by prior work. We summarize the representative insights in the following.

\input{tables/auroc_main}

\vsp
\paragraph{New findings.} Our experiments using \benchmarkname provide new insights:
\begin{enumerate}
\item \benchmarkname spans a wide range of difficulty levels, enabling effective evaluation of hallucination detectors.
\item It reveals performance differences across models that are not visible in existing benchmarks, including MMMU and prior alignment benchmarks.
\item Detectors exhibit an \textit{early-sentence bias}, tending to judge sentences appearing early in a response as \textit{correct}.
\item Dataset noise can be reduced by filtering captions with strong VLM detectors, regardless of whether the captions are generated by strong or weak VLM captioners.
\end{enumerate}

\vsp
\paragraph{Confirmed findings.} We yielded consistent results with several known insights:
\begin{enumerate}
   \item Detectors exhibit strong self-preference, favoring captions generated by the same model.
    \item CLIP-like models struggle to align pairs generated by modern captioners.
    \item Even advanced VLMs exhibit weakness in detecting errors in counting the number of objects. 
\end{enumerate}

\subsection{Overview of Results}
Table~\ref{tab:auc_compare} presents the results evaluated on diverse VLMs using Captioners as evaluation targets. Samples of detectors' outputs are available in Fig.~\ref{fig:ex_scoring}.

\nbf{\benchmarkname covers diverse levels of hallucination detection.} We see variations in the performance across the tested VLMs and caption models. Thus, \benchmarkname is suitable to quantify VLMs as hallucination detection models.
\input{figures/ex_scoring}

\input{figures/mmmu_vs_auroc}
\input{figures/positionwise_scores}
\nbf{CLIP-based models tailored for compositional alignment remain nearly blind.}
TripletCLIP~\cite{patel2024tripletclip}, despite being trained for compositional understanding, achieves an AUROC of only around 50, indicating that it fails to distinguish correct from incorrect sentences. The same trend holds BLIP-2.

\nbf{Best model.} On average, Gemini-3-Pro shows the best performance of all models, while Llama-4, the best open-source model, performs on par with GPT-5-mini. Llama-4 outperforms many private models with a large margin. Its activated parameters during inference are only 17B. When considering the balance of inference time and accuracy, Llama-4 is the best in open-source models.

\nbf{Examples of detectors' outputs.} Figure~\ref{fig:ex_scoring} illustrates that even advanced detectors can misinterpret the visual content, despite the captioners’ errors not being particularly subtle, as in the center-left example.

\input{figures/likelihood_analysis}
\input{tables/compare_other_dataset}

\subsection{Detailed Analysis}
\label{sec:detailed_analysis}
We here describe more detailed insights from the experiments.

\nbf{Advanced captioners produces hard-to-detect errors.} Hallucinations from GPT-4o, Gemini-3, and GPT-5 are difficult to detect, even for proprietary models, as indicated by their relatively low detection scores. Since these captioners accurately understand many scenes, their errors tend to be subtle and therefore harder to identify. For example, Gemini-3 produces errors in 6.5\% of sentences, whereas ShareGPT produces errors in 26.5\% of sentences.
The left of Fig.~\ref{fig:two_figs} plots captioner performance on MMMU (x-axis) against AUROC measured by GPT-5-mini (y-axis). Captioners with higher MMMU scores tend to yield lower AUROC, indicating that stronger captioners generate hallucinations that are harder to detect. 
\looseness=-1

\nbf{\benchmarkname measures visual understanding abilities that are not captured by existing benchmarks.} The right of Fig.~\ref{fig:two_figs} plots the performance on MMMU (x-axis) and \benchmarkname (y-axis). Although the performance on the two benchmarks is correlated overall, better performance on MMMU does not necessarily ensure better performance on \benchmarkname. For instance, Qwen-3-VL-235B and Qwen-2.5-VL-32B outperform Llama-4 on MMMU, while Llama-4 outperforms them on \benchmarkname. This result indicates that evaluating on the image reasoning benchmark does not necessarily reveal the fine-detailed visual understanding required for hallucination detection. 

\nbf{Detectors favor the sentence near the beginning of the caption.} The left of Fig.~\ref{fig:positionwise_score} shows the detectors' output scores averaged across each sentence position. For both \textit{correct} and \textit{incorrect} image-text pairs, the detectors assign higher scores to sentences located near the beginning of the caption.
The right of Fig.~\ref{fig:positionwise_score} shows the likelihood of generating each sentence given the image across sentence positions, measured by the detector. The first sentence tends to receive higher likelihood scores, suggesting that it often contains more generic descriptions that provide an overview of the image. VLMs appear to prefer such sentences irrespective of their correctness, possibly because such patterns are abundant in training data.

\nbf{Detectors prefers the hallucinated sentences in terms of likelihood.}
We investigate the behavior of the language model within a VLM when processing hallucinated and non-hallucinated sentences. Specifically, we measure the likelihood produced by the decoder (LLama-4 400B) for a given sentence while pairing it with a randomly sampled image, in order to minimize the influence of the input image. Figure~\ref{fig:likelihood_analysis} shows the likelihood distribution across sentence positions. Surprisingly, the decoder assigns higher likelihood to \textit{hallucinated} sentences, suggesting that these sentences are more plausible from the perspective of the language model's learned language patterns. We observe a similar trend even when the sentence is paired with its corresponding image. Previous work~\cite{suo2025octopus,jiang2025devils,favero2024multi} has shown that hallucinations often arise when the captioner disregards visual signals and instead generates tokens that are plausible under language modeling. Such linguistically plausible tokens may also be favored by detector VLMs. 
Interestingly, despite this tendency, LLama-4 still achieves strong performance on ShareGPT and LLaVA (Table~\ref{tab:auc_compare}). This result suggests that the detector does not rely solely on sentence likelihood when judging caption correctness.

\nbf{\benchmarkname can serve as a new indicator of an image-text alignment capability.}
Table~\ref{tab:compare_other_dataset} compares \benchmarkname with existing benchmarks for hallucination detection and vision–language compositionality. Strong performance on prior benchmarks does not necessarily translate to strong performance on \benchmarkname. For example, although Gemma-3 achieves strong results on Winoground, FOIL, and HAT, its performance drops substantially on \benchmarkname, suggesting that success on short and relatively simple sentences does not generalize to the longer and more diverse captions in \benchmarkname. In contrast, Llama-4 performs well on \benchmarkname while also achieving strong results on existing benchmarks, yet the performance drops on Winoground and HAT, which involve complex object compositions. Overall, these results suggest that evaluation on both \benchmarkname and existing benchmarks is necessary to develop robust hallucination detectors.

\input{figures/noise_reduction_performance}
\nbf{VLM-based image-sentence filtering can reduce hallucinated image-sentence pairs.}
We evaluate the effectiveness of using VLMs to clean image-sentence paired datasets.
Figure~\ref{fig:noise_reduction_performance} measures the ability to remove unaligned image-sentence pairs when Llama-4 (109B) is used as a detector.
Given the alignment scores, we select the top 20\% of sentences and measure their precision. The noise rate substantially decreases across all caption sources.
The improvement is particularly pronounced for weaker captioners such as ShareGPT, while the noise rate for the performance Captioner, GPT-4o, is also reduced to 5.0\%. Llama-4's AUROC on GPT-4o is only 67.8, yet picking the confident examples can effectively reduce the noise rate. In the appendix, we present additional experiments on filtering when learning from VLM-generated captions and confirm its effectiveness.

\input{figures/hal_category_scoring}

\nbf{Detectors are poor at detecting \textit{Direction} and \textit{Number} hallucinations.} 
Figure~\ref{fig:hal_category_analysis} assesses detectors’ robustness across hallucination categories using their output scores (lower is better since only hallucinated sentences are accounted). \textit{Direction} errors occur when object orientation is misdescribed; identifying the errors requires fine-grained visual understanding, and detectors consistently perform poorly. \textit{Number} errors arise from incorrect object counts—an issue long recognized in early VLMs like CLIP~\cite{paiss2023teaching} and still evident in advanced models.

\input{tables/t2i_result}
\nbf{Results on hallucination detection for images generated by text-to-image models.}
Table~\ref{tab:t2i_auc_compare} presents results for hallucination detection in image-text pairs using images generated by diverse text-to-image models. We have two observations: (i) stable diffusion produces errors easy to detect, while errors of many proprietary models are hard to detect by strong models, (ii) GPT5 consistently outperforms other models in all cases. These results and those of Table~\ref{tab:auc_compare} conclude that GPT5 is a strong hallucination detector for image-text pairs in many cases.

\input{tables/localization_miou_ap}

\nbf{Segment-level localization has more room for improvement.} 
\benchmarkname includes hallucination segments for each hallucinated sentence, enabling segment-level evaluation. We present VLMs with a hallucinated sentence-image pair and prompt them to localize the hallucinated span, explicitly noting that one exists. Performance is measured by alignment with human annotations (see Appendix for prompts and metrics). As shown in Table~\ref{tab:localization_combined}, Llama-4 (400B), the best model, localizes only 24.2\% of hallucinated segments on average, underscoring substantial room for improvement. Notably, GPT-4o mini outperforms Qwen-2.5 (32B), in contrast to Table~\ref{tab:auc_compare}, indicating that strong sentence-level detectors are not always effective for segment-level localization.

\input{figures/self_preference_cvpr}

\nbf{Detectors struggle to detect their own hallucinations.}
Table~\ref{tab:auc_compare} shows that Llama-4 (109B), GPT-4o, and Gemini-3-Pro perform poorly on their own outputs (highlighted by underline). 
This finding aligns with prior work reporting LLM detectors favor their own outputs~\cite{panickssery2024llm}. We annotate captions generated by Qwen-2.5 (32B) and Gemma-3 (27B) to enable more extensive self- and cross-evaluations. Figure~\ref{fig:self_pref_score} (left) confirms much lower AUROC on self-generated captions (diagonal elements). Figure~\ref{fig:self_pref_score} (right) shows that GPT-4o scores its own \textit{incorrect} sentences higher than those of Llama-4, and the gap between incorrect and correct scores is small in its own output, which is causing the performance degradation. 
\looseness=-1

\input{tables/ensemble}

\nbf{Ensembling improves performance.} We study whether ensembling improves detection. We average alignment scores from two comparably strong models (Table~\ref{tab:auc_compare}) and observe gains in most cases (Table~\ref{tab:ensemble}). This suggests that models apply distinct criteria for image-caption alignment, and combining them enhances performance. A drop occurs for ensembling Llama-4 and GPT-5-mini on Llama-4 captions, likely due to the large performance gap between the two models. The improvements on Gemini-3 are limited, which indicates that boosting each detector's performance is crucial in detecting hallucinations in challenging cases.

%% file: tables/auroc_main.tex
\begin{table*}[t!]
  \small
  \centering
  \caption{\small AUROC results across VLMs.}
  \vspace{-2mm}
  \fontsize{6.2pt}{9.0pt}\selectfont
  \setlength\tabcolsep{3.0pt}
  \renewcommand{\arraystretch}{0.95}

  \scalebox{0.95}{
  \begin{tabular}{lc|cc|ccc|ccc|c}
  \toprule
  \multirow{4}{*}{\textbf{Detector}} &\multirow{4}{*}{\textbf{Size}} &

  \multicolumn{8}{c|}{\textbf{Captioners}} \\
\cmidrule(lr){3-4}
\cmidrule(lr){5-7}
\cmidrule(lr){8-10}
  & &\textbf{S-GPT} & \textbf{LLaVA} & \textbf{Qwen} & \textbf{CogV}
  & \textbf{Llama-4} & \textbf{GPT4o} &
  \textbf{Gemini-3} & \textbf{GPT5}   &\textbf{Avg.} \\
  \hline
  \rowcolor{gray!11}
  \multicolumn{11}{l}{\textbf{Open-Source}} \\
  BLIP~\cite{li2022blip} &1.2B& 53.5&55.9&52.4&52.1&52.2&52.5&52.0&50.1&52.6\\
  TripletCLIP~\cite{patel2024tripletclip} &0.3B& 50.7&51.9&54.0&51.0&50.9&53.9&51.3&52.6&52.0\\
  Qwen-2~\cite{wang2024qwen2}&7B & 60.2&55.3&55.9&51.9&53.2&46.0&49.8&53.3&53.2\\
  LLaVA-Next~\cite{li2024llavanext-strong} &72B& 59.4&56.6&58.9&57.5&55.0&56.7&53.0&50.2&55.9\\
  Pixtral~\cite{agrawal2024pixtral}&12B& 64.3&60.8&60.6&57.3&55.7&57.1&51.7&57.5&58.1\\
  \arrayrulecolor{gray!40}\cline{1-11}\arrayrulecolor{black}
  \multirow{2}{*}{Gemma~\cite{gemmateam2025gemma3technicalreport}} &12B& 67.0&63.3&63.1&59.4&54.0&55.7&53.3&61.8&59.7\\
  &27B& 67.6&64.4&67.9&66.4&60.6&61.1&54.8&69.6&64.0\\
  \arrayrulecolor{gray!40}\cline{1-11}\arrayrulecolor{black}
\multirow{3}{*}{InternVL2~\cite{chen2024far}} &8B & 66.6&65.7&69.1&67.6&60.6&63.9&55.9&43.9&61.7\\
  &26B& 63.9&60.6&61.2&58.9&55.6&57.1&52.1&57.0&58.3\\
  & 40B & 69.3&63.0&66.5&61.9&61.5&59.6&58.5&53.8&61.8\\
  \arrayrulecolor{gray!40}\cline{1-11}\arrayrulecolor{black}
  \multirow{2}{*}{Qwen-2.5~\cite{bai2025qwen2}} &7B & 68.6&65.3&66.5&64.6&61.0&55.7&56.5&61.3&62.4\\
  & 32B & 73.6&71.6&70.6&69.0&\textbf{66.0}&66.1&52.9&61.9&66.5\\
  \arrayrulecolor{gray!40}\cline{1-11}\arrayrulecolor{black}
  \multirow{2}{*}{Qwen3~\cite{bai2025qwen3}} & 30B& 79.3 & 74.8 & 75.8 & 72.2 &64.8& 66.0 &\textbf{62.2}& 50.4&68.2\\
  &235B& \textbf{81.9}&76.1&76.8&73.4&65.2&65.5&54.2&57.6&68.8\\
  \arrayrulecolor{gray!40}\cline{1-11}\arrayrulecolor{black}
  \multirow{2}{*}{Llama-4~\cite{llama4}}  &108B & 80.8&78.7&78.2&77.4&\underline{59.9}&67.8&56.2&\textbf{65.8}&70.6\\
  &400B & 81.1&\textbf{80.9}&\textbf{79.0}&\textbf{81.3}&64.7&\textbf{71.9}&53.0&65.1&\textbf{72.1}\\
  \hline
  \rowcolor{gray!11}
  \multicolumn{11}{l}{\textbf{Closed Models}} \\
  Gemini-2.0-flash &N/A& 76.8&73.5&74.9&70.4&64.8&65.5&55.9&62.9&68.1\\ 
  Gemini-3-Pro &N/A& 85.2 & \textbf{86.2}&\textbf{86.2}& \textbf{87.4}&\textbf{83.0}&\textbf{80.7}&\underline{54.2}&\textbf{75.8}&\textbf{79.8}\\
   \arrayrulecolor{gray!40}\cline{1-11}\arrayrulecolor{black}
  GPT4o &N/A& 75.8&72.6&72.8&69.7&63.9&\underline{58.2}&58.9&55.9&65.9\\
  GPT5-mini& N/A& 81.5&82.2&80.2&81.1&73.0&69.7&55.1&64.1&73.4\\
  GPT5 & N/A&\textbf{85.4}&86.0&85.3&85.5&78.4&72.3&\textbf{60.4}&\underline{69.8}&77.9\\

  \bottomrule
  \end{tabular}
  }
  \label{tab:auc_compare}
\end{table*}

%% file: figures/ex_scoring.tex
\begin{figure*}[t]
\centering
\includegraphics[width=\linewidth]{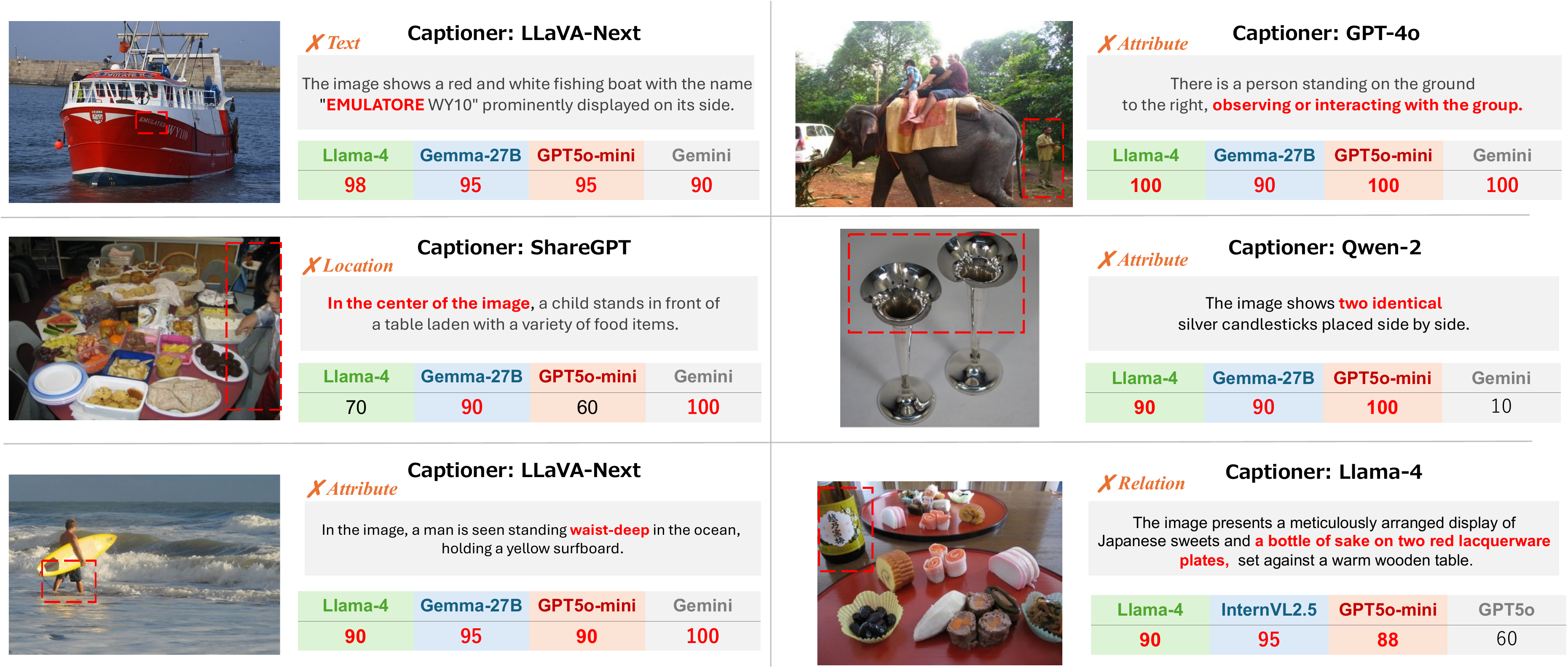}
\vspace{-5mm}
\caption{\small \textbf{Examples of \textit{incorrect} sentences with detectors’ correctness scores.} Higher scores indicate greater confidence in correctness. Detectors are prone to being overconfident in these examples. We highlight detectors’ errors in red within the text and mark the grounded \textit{incorrect} regions in the image with orange boxes.}
\label{fig:ex_scoring}
\end{figure*}

%% file: figures/mmmu_vs_auroc.tex
\begin{figure*}[t]
\centering
\begin{subfigure}{0.49\linewidth}
  \centering
  \includegraphics[width=0.95\linewidth]{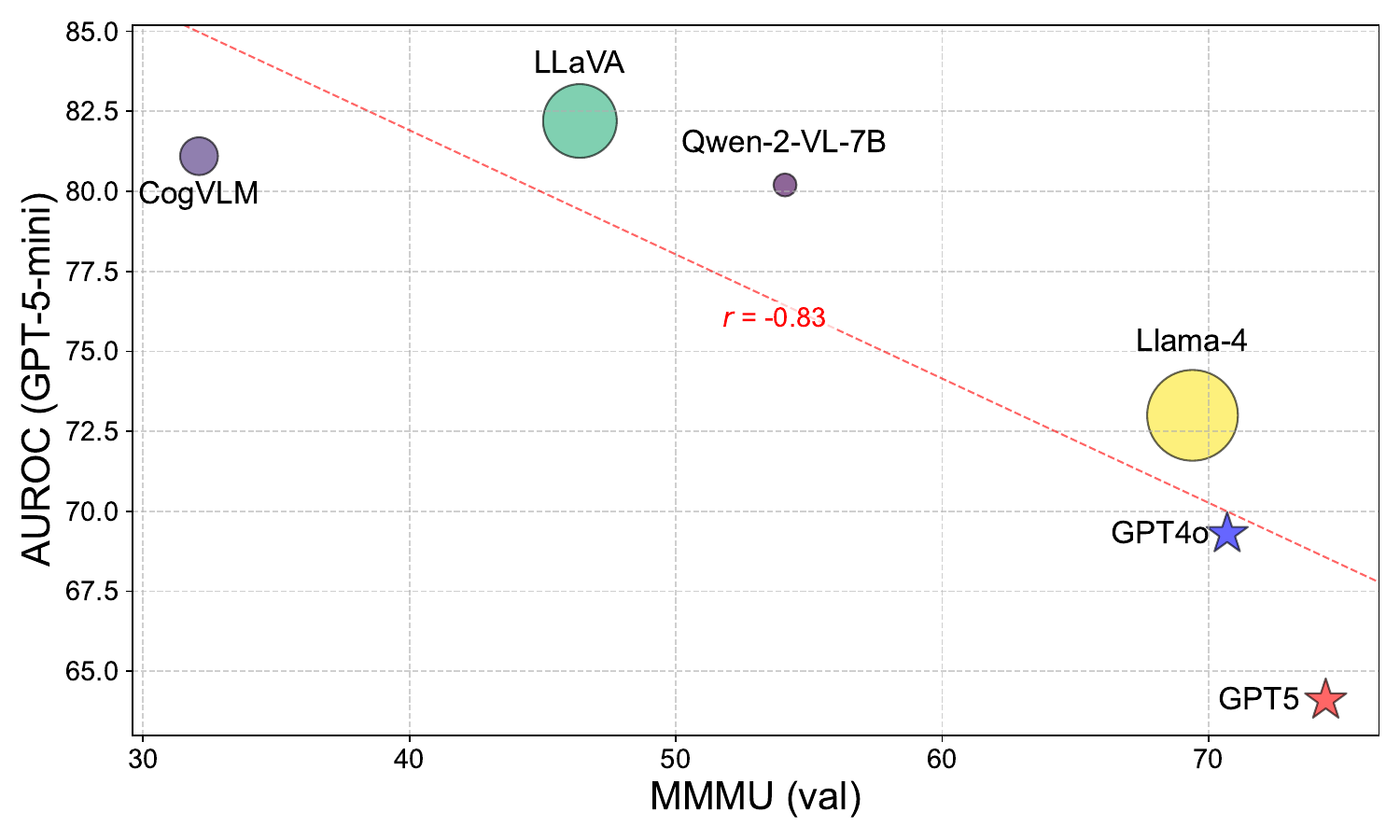}
  \label{fig:mmmu_vs_auroc_captioners}
\end{subfigure}
\hfill
\begin{subfigure}{0.49\linewidth}
  \centering
  \includegraphics[width=0.95\linewidth]{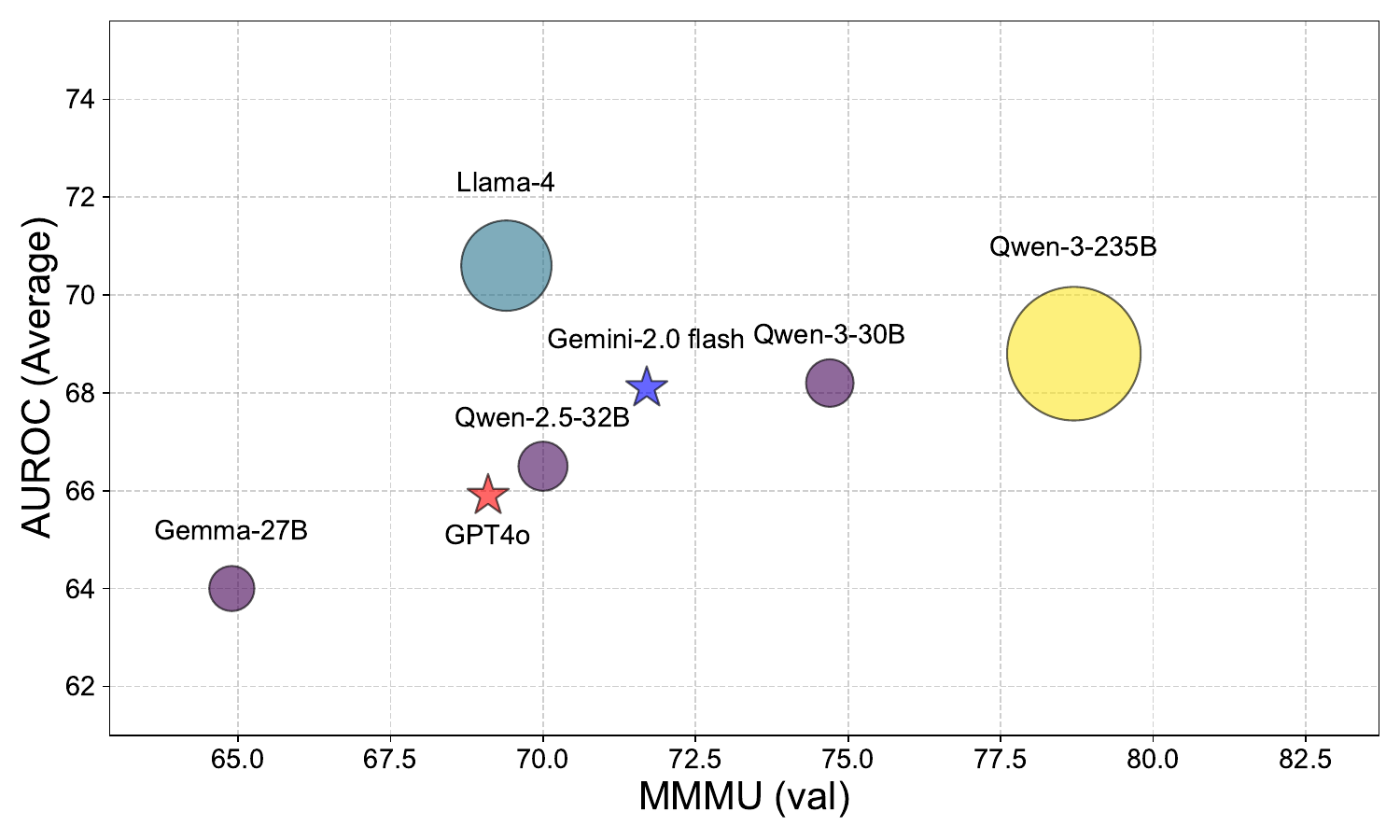}
  \label{fig:mmmu_vs_auroc_detectors}
\end{subfigure}
\vspace{-3mm}
\caption{\small The size of plots indicates the parameter size. \textbf{Left:} MMMU performance measured on \textit{Captioners} (X-axis) vs. AUROC measured by GPT-5-mini (Y-axis) for each Captioner. Advanced Captioners tend to produce errors that are difficult to detect. \textbf{Right:} MMMU (X-axis) vs. AUROC (Y-axis) for each \textit{detector}. Detectors with better MMMU performance do not necessarily perform better on \benchmarkname. }
\label{fig:two_figs}
\end{figure*}

%% file: figures/positionwise_scores.tex
\begin{figure*}[t]
\centering
\includegraphics[width=\linewidth]{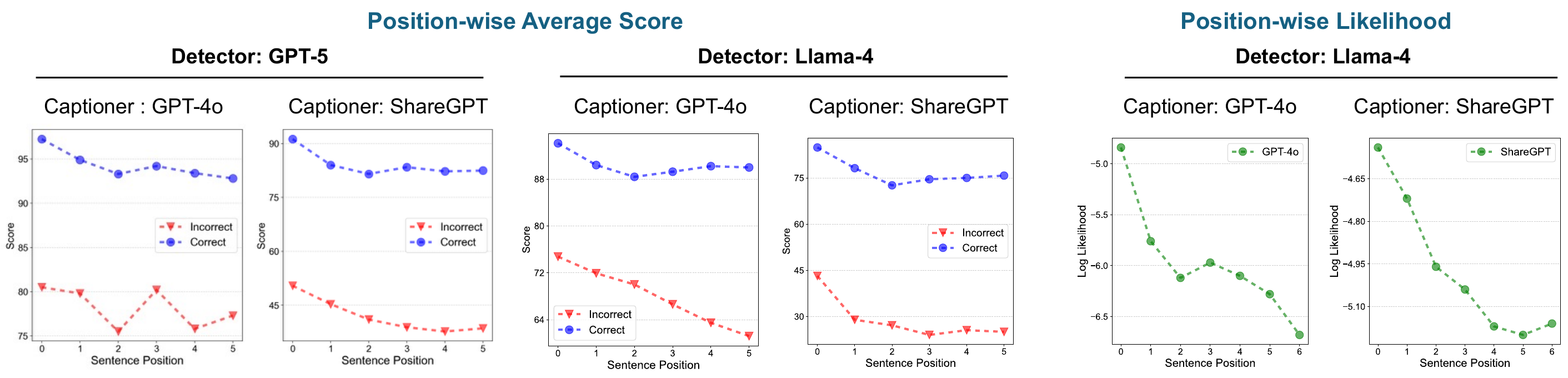}
\vspace{-5mm}
\caption{\small \textbf{Detectors show positional bias in scoring.} We average the detectors’ correctness scores (Y-axis) by sentence position (X-axis) and visualize the results using GPT-4o (Left) and Llama-4 (Right) as detectors. Both detectors assign higher scores to sentences appearing near the beginning of the output. The detector is \textit{not} provided with any positional information during inference.}
\label{fig:positionwise_score}
\end{figure*}

%% file: figures/likelihood_analysis.tex
\begin{figure}[t]
\centering
\includegraphics[width=0.9\linewidth]{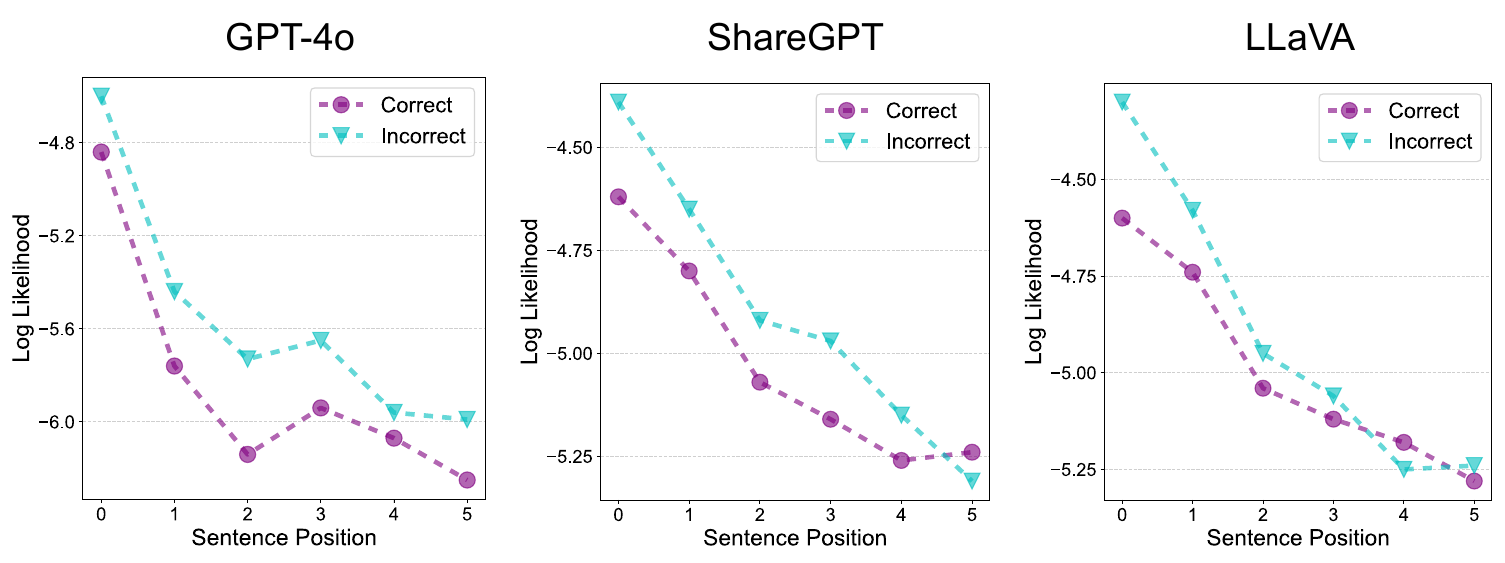}
\vspace{-3mm}
\caption{\small The likelihood of sentences measured by the detector, Llama-4. Incorrect sentences tend to be more plausible than correct ones in terms of the likelihood.}
\label{fig:likelihood_analysis}
\end{figure}

%% file: tables/compare_other_dataset.tex
\begin{table*}[t]
  \small
 \caption{\small \textbf{Comparison to existing datasets for hallucination detection and compositionality understanding} (AUROC). 
$\uparrow$ indicates that the result on the prior benchmark exceeds the performance on \benchmarkname.}
  \vspace{-3mm}
  \centering
    \fontsize{6.2pt}{8.0pt}\selectfont
    \setlength\tabcolsep{4.0pt} 
  \newcommand{\up}{\hspace{0.15em}{\color{blue!70!black}$\uparrow$}}
    \renewcommand{\arraystretch}{1.2} 
  \resizebox{\linewidth}{!}{
  \begin{tabular}{cc!{\vrule width 0.6pt}cccc
                c!{\vrule width 0.6pt}ccc
                c!{\vrule width 0.6pt}cccc}
    \toprule
   \multirow[c]{2}{*}{{\textbf{Detector}}}  &&
    \multicolumn{4}{c}{\textbf{\benchmarkname}} &&
    \multicolumn{3}{c}{\textbf{Hallucination Detection}} & & \multicolumn{4}{c}{\textbf{VL-Compositionality}} \\
    \cmidrule(lr){3-6} \cmidrule(lr){8-10}  \cmidrule(lr){12-15}
     & &\textbf{S-GPT}& \textbf{GPT4o} &
    \textbf{Llama} &\textbf{CogV}& & \textbf{MHalD} & \textbf{Foil} &\textbf{HAT}&& \textbf{ARO}& \textbf{SugarCrp} & \textbf{Winogrnd}&\textbf{SeeTrue}\\\hline
    Qwen-2.5 (32B) && 73.6&66.1&66.0&69.0&&83.0\up&89.6\up&77.6\up&&78.3\up&84.9\up&73.1&81.5\up\\
Gemma-3 (27B) && 67.6&61.1&60.6&66.4&&81.7\up&91.8\up&76.6\up&&79.2\up&87.6\up&77.6\up&81.5\up\\
Llama-4 (109B) && 80.8&67.8&59.9&77.2&&82.8\up&90.3\up&76.3&&84.8\up&89.4\up&77.1&83.1\up\\
GPT-5-mini &&81.5&69.7&73.0&81.1&&85.4\up&94.6\up&84.6\up&&92.4\up&92.4\up&90.8\up&87.2\up\\

    \bottomrule
    \end{tabular}
  } 
  \label{tab:compare_other_dataset}
\end{table*}

%% file: figures/noise_reduction_performance.tex
\begin{figure}[t]
\centering
\includegraphics[width=\linewidth]{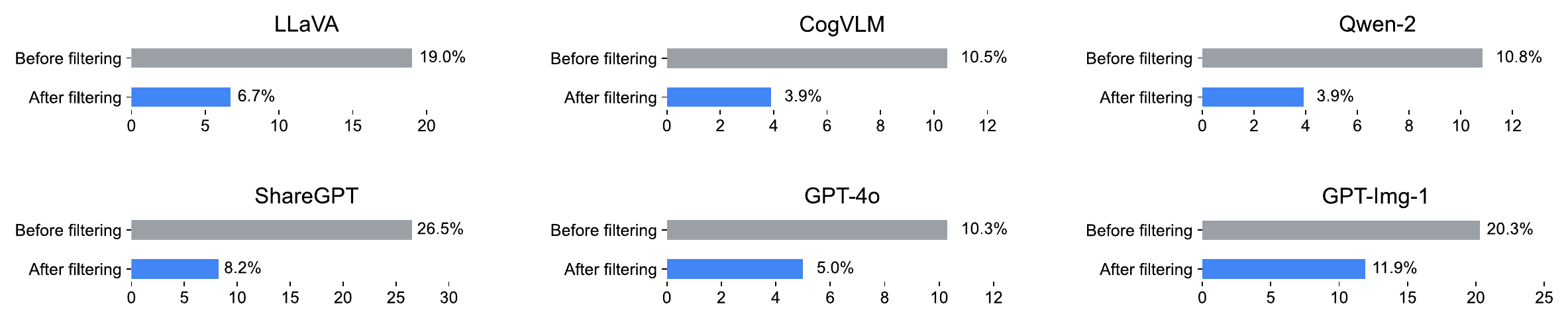}
\vspace{-5mm}
\caption{\small The performance of image-caption pair filtering using Llama-4. We pick image-sentence pairs with top-20\% alignment score, which effectively reduces the ratio of unaligned image-sentence pairs.}
\label{fig:noise_reduction_performance}
\end{figure}

%% file: figures/hal_category_scoring.tex
\begin{figure}[t]
\centering
\includegraphics[width=\linewidth]{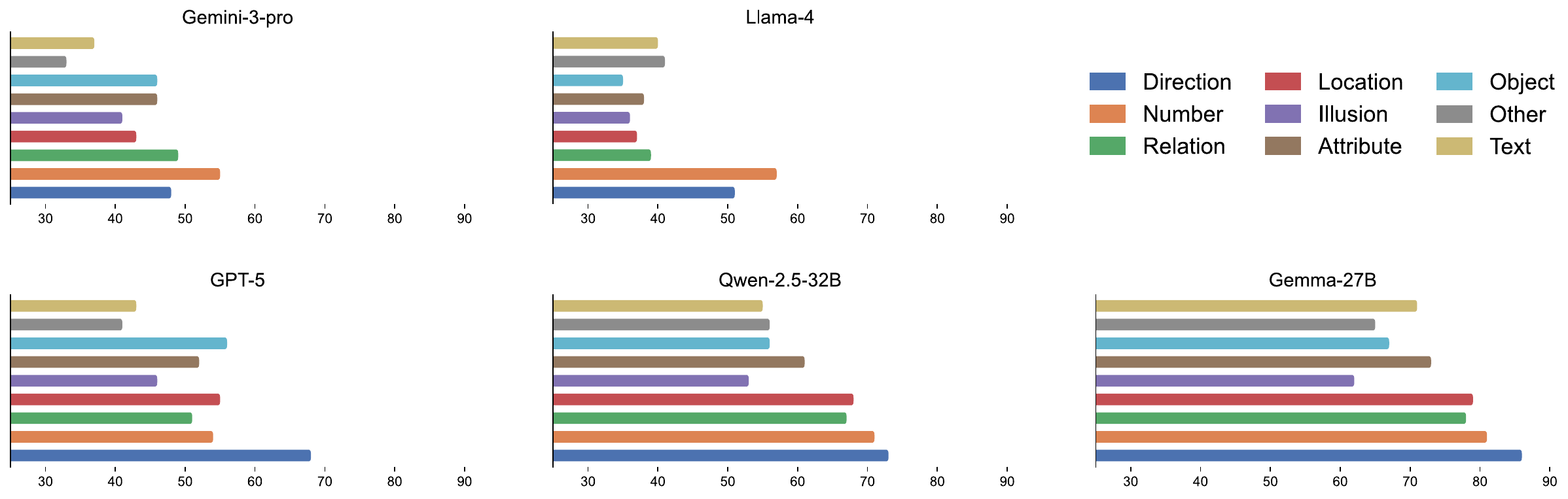}
\caption{\small \textbf{Detectors' score averaged within each hallucination type} (lower is better.) Detectors show weakness in \textit{Direction} (blue) and \textit{Number} (orange).}
\label{fig:hal_category_analysis}
\end{figure}

%% file: tables/t2i_result.tex
\begin{table*}[t!]
  \small
  \centering
  \caption{\small Hallucination detection results on image-text pairs generated using Text-to-Image models (AUROC).}
  \vspace{-2mm}
  \fontsize{6.2pt}{9.0pt}\selectfont
  \setlength\tabcolsep{5.0pt}
  \renewcommand{\arraystretch}{0.95}
  \resizebox{\linewidth}{!}{
  \begin{tabular}{l|ccccc|c}
  \toprule
  \multirow{4}{*}{\textbf{Detector}} &
  \multicolumn{5}{c|}{\textbf{Text-to-Image models}} \\
  \cmidrule(lr){2-6}
  & \textbf{Stable Diffusion}
  & \textbf{GPT-Img-1}
  & \textbf{GPT-Img-1.5}
  & \textbf{Gemini-2.5}
  & \textbf{Gemini-3}
  & \textbf{Avg.} \\
  \hline
  Qwen-2-7B & 54.6 & 46.0 & 57.2 & 63.8 & 59.8 & 56.3 \\
  Qwen-2.5-32B & 68.9 & 61.0 & 59.4 & 65.0 & 62.6 & 63.4 \\
  Gemma-27B & 63.7 & 50.5 & 52.8 & 57.7 & 53.6 & 55.7 \\
  Qwen3-235B & 83.0 & 66.4 & 65.1 & 67.3 & 67.1 & 66.5 \\
  Llama-4-400B & 83.0 & 67.8 & 64.3 & 66.8 & 65.4 & 69.5 \\
  GPT5-mini & 83.8 & 65.7 & 64.8 & 65.7 & 66.3 & 69.3 \\
  GPT5 &\textbf{84.9} & \textbf{72.0} &\textbf{70.0} & \textbf{71.9} & \textbf{67.4} & \textbf{73.2} \\

  \bottomrule
  \end{tabular}
  }
  \label{tab:t2i_auc_compare}
\end{table*}

%% file: tables/localization_miou_ap.tex
\begin{table*}[t]
  \centering
   \caption{\small \textbf{Hallucinated segment localization results.} Each cell shows AP / mIoU (\%). Localizing hallucinated segments remains difficult even for performant models.} 
    \fontsize{7.2pt}{8.0pt}\selectfont
    \setlength\tabcolsep{5.0pt}
    \renewcommand{\arraystretch}{1.5}
  \resizebox{\linewidth}{!}{
    \begin{tabular}{ccccccccccccc}
    \toprule
   \multirow[c]{2}{*}{{\textbf{Detector}}} &  
   \multirow[c]{2}{*}{{\textbf{Params}}} & &
    \multicolumn{6}{c}{\textbf{Image-to-Caption Models}} & &
    \multicolumn{2}{c}{\textbf{Text-to-Image Models}} &
    \multirow[c]{2}{*}{{\textbf{Avg.}}} \\
    \cmidrule(lr){4-9} \cmidrule(lr){11-12}
     & & & \textbf{S-GPT} & \textbf{Llava} & \textbf{Qwen-2} & \textbf{GPT4o} &
    \textbf{CogVLM} & \textbf{Llama-4} && \textbf{Stable Diffusion} & \textbf{GPT-Img-1} \\\hline    

    Qwen-2.5 & 32B & &
    17.3 / 13.8 &
    22.4 / 15.1 &
    14.0 / 11.7 &
    20.8 / 15.1 &
    22.5 / 16.0 &
    12.6 / 10.7 && 
    \textbf{15.9} / 11.4 &
    12.1 / 9.4 &
    17.2 / 12.9 \\

    GPT-4o mini & - & &
    25.2 / 21.6 &
    28.9 / \textbf{22.4} &
    20.6 / 18.3 &
    29.5 / 23.3 &
    28.0 / 21.5 &
    18.7 / 16.4 && 
    14.7 / \textbf{12.5} &
    \textbf{14.6} / \textbf{11.9} &
    22.5 / 18.5 \\

    Llama-4 & 109B & &
    24.9 / 22.6 &
    27.0 / 20.8 &
    24.8 / 22.7 &
    \textbf{34.3} / \textbf{26.4} &
    29.3 / \textbf{23.2} &
    \textbf{19.6} / 17.3 && 
    15.1 / 10.7 &
    11.0 / 9.0 &
    23.3 / 19.1 \\

    Llama-4 & 400B & &
    \textbf{28.2} / \textbf{24.8} &
    \textbf{29.1} / 22.1 &
    \textbf{26.2} / \textbf{23.3} &
    34.0 / 26.0 &
    \textbf{29.8} / 21.7 &
    19.4 / \textbf{18.0} && 
    15.4 / 11.9 &
    12.0 / 9.3 &
    \textbf{24.2} / \textbf{19.6} \\   

    \bottomrule
    \end{tabular}
  }   
  \label{tab:localization_combined}
\end{table*}

%% file: figures/self_preference_cvpr.tex
\begin{figure}[t]
\centering
\includegraphics[width=\linewidth]{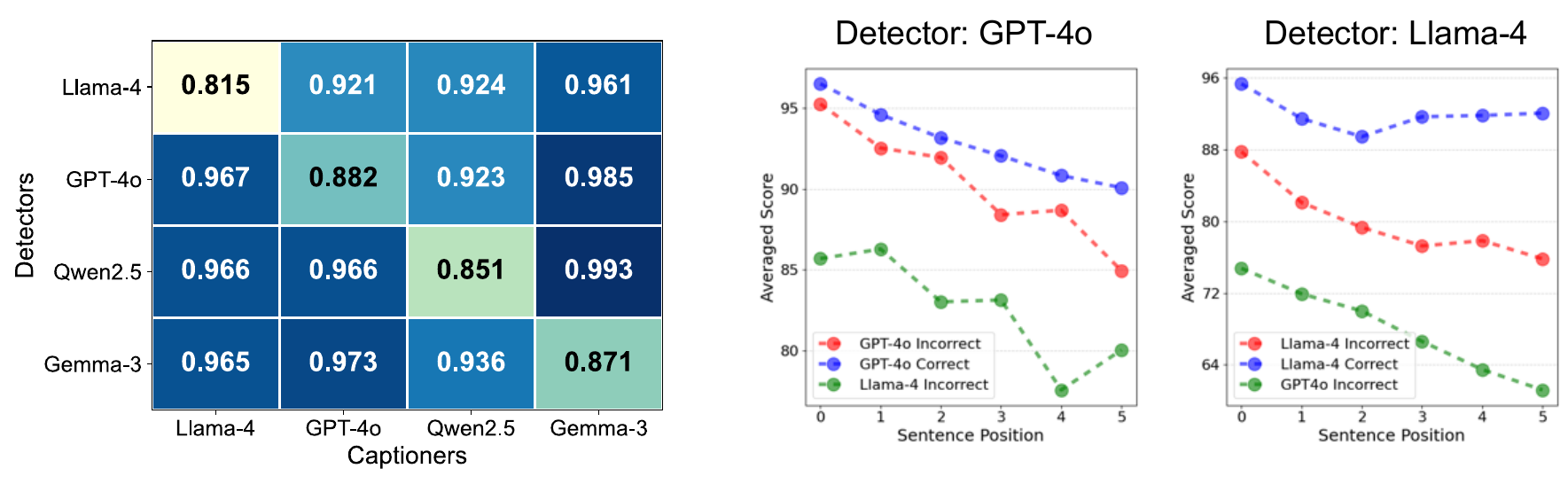}
\vspace{-6mm}
\caption{\small \textbf{Detectors struggle to detect their own hallucination.} \textbf{Left:} Self- and cross-evaluation results. AUROC scores for each Captioner (columns), normalized by the average AUROC of each Detector (rows). Diagonal entries show self-evaluation. \textbf{Right:} We pick GPT-4o as a detector, with their output correctness scores averaged by sentence position. \textcolor{blue}{Blue} and \textcolor{red}{red} lines show scores for \textit{correct} and \textit{incorrect} GPT-4o's outputs; \textcolor[HTML]{006400}{green} shows scores for \textit{incorrect} Llama-4 outputs.}
\label{fig:self_pref_score}
\end{figure}

%% file: tables/ensemble.tex
\begin{table*}[h!]
  \centering
  \caption{\small \textbf{Ensembling improves performance in many cases.} 
  The performance difference from the \textit{better} model used for ensembling is highlighted.}
  \vspace{-2mm}
  \fontsize{6.2pt}{8.0pt}\selectfont
    \setlength\tabcolsep{3.0pt} 
  \renewcommand{\arraystretch}{2.0}
  \newcommand{\plusvalue}[1]{{\textcolor{blue}{\fontsize{2.5pt}{2.5pt}\selectfont (+#1)}}}
\newcommand{\minusvalue}[1]{{\textcolor{red}{\fontsize{2.5pt}{2.5pt}\selectfont (-#1)}}}
\resizebox{\linewidth}{!}{
  \begin{tabular}{c|cccccccc}
  \toprule
  \textbf{Detectors} &
  \textbf{S-GPT} &
  \textbf{LLaVA} &
  \textbf{Qwen-2} &
  \textbf{GPT4o} &
  \textbf{CogVLM} &
  \textbf{Llama-4} &
  \textbf{GPT-5} &
  \textbf{Gemini-3} \\
  \hline

  \makecell{Llama-4 (109B) \\ Llama-4 (400B)}
  & 84.6\plusvalue{3.5}
  & 83.4\plusvalue{2.4}
  & 82.9\plusvalue{3.9}
  & 74.0\plusvalue{2.1}
  & 83.5\plusvalue{2.2}
  & 65.8\plusvalue{1.1}
  & 68.7\plusvalue{2.9}
  & 56.2\plusvalue{0.0} \\
 \arrayrulecolor{gray!40}\cline{1-9}\arrayrulecolor{black}
  \makecell{Llama-4 (109B) \\ GPT5-mini}
  & 86.0\plusvalue{4.5}
  & 84.8\plusvalue{2.7}
  & 83.6\plusvalue{3.4}
  & 73.9\plusvalue{4.2}
  & 84.4\plusvalue{3.2}
  & 72.3\minusvalue{0.7}
  & 68.7\plusvalue{2.9}
  & 57.9\plusvalue{1.7} \\

  \bottomrule
  \end{tabular}}

  \label{tab:ensemble}
\end{table*}

%% file: chapter/5_conclusion.tex
\section{Conclusion}
\vspace{-2mm}
We introduced \benchmarkname, a benchmark for evaluating hallucination detection in image captions (HalDec). \benchmarkname enables systematic evaluation of detectors across different caption styles, hallucination types. Our experiments reveal that strong performance on existing benchmarks does not necessarily translate to strong performance on \benchmarkname, highlighting limitations of current evaluation settings. Furthermore, our analysis uncovers several empirical findings, including a positional bias in detector judgments and the effectiveness of using strong VLMs as filters to reduce dataset noise. We hope that \benchmarkname will facilitate more reliable evaluation of hallucination detectors and contribute to improving image-text alignment in future VLMs.

%% file: supplemental_arxiv.tex
 



\newcommand{\plusvalue}[1]{\hspace{0.3em}\textcolor{teal}{{\fontsize{7pt}{8pt}\selectfont\textbf{(+#1)}}}}
\newcommand{\minusvalue}[1]{\hspace{0.3em}\textcolor{red}{{\fontsize{7pt}{8pt}\selectfont\textbf{(-#1)}}}}

\definecolor{cvprblue}{rgb}{0.21,0.49,0.74}

\definecolor{lightcyan}{rgb}{0.88,0.95,1}

\newcommand{\colordmark}{{\ding{52}}}
\newcommand{\colorxmark}{{\ding{55}}} 
\sloppy

\renewcommand{\thesection}{\Alph{section}}
\renewcommand{\thetable}{\Alph{table}}
\renewcommand{\thefigure}{\Alph{figure}}
\setcounter{figure}{0}
\setcounter{table}{0}
\section{Limitation}

\noindent\textbf{Methodology.}
HalDec needs to be a light-weight model, considering its application to curate datasets. However, our results indicate that VLMs with more parameters show superior performance. Also, our evaluation relies on sentence-by-sentence score output, which regards each sentence as independent. However, this protocol ignores the context of consecutive sentences. We observe that many sentences can be regarded as independent, yet considering multiple sentences together might improve the performance of hallucination detection. 

\noindent\textbf{Annotations.}
Judging the hallucinations in image captions involves subjective criteria of annotators. Captions may look hallucinated to some annotators, while they do not to others. Having a unified consensus on this criterion is difficult. For sentence-level annotation, we introduce a category \textit{unknown}, which allows us to exclude such ambiguous samples during evaluation. This issue can be more significant in segment localization and categorizing hallucination types. Then, we focus on sentence-level detection to benchmark VLMs following Mishra et al. ~\cite{refA}.

\section{The Use of Large Language Models (LLMs)}
In preparing this manuscript, we made limited use of large language models (LLMs) such as ChatGPT. Specifically, LLMs were employed only to assist with polishing the writing for grammar, clarity, and readability. No part of the research design, analysis, interpretation, or results was generated or influenced by LLMs. All scientific content, data, and conclusions are the sole work of the authors.

\section{Attribution to an icon}
We employ the chatbot icons created by Freepik - Flaticon\footnote{https://www.flaticon.com/free-icons/chatbot} for Fig.~\ref{fig:teaser}.

\section{Dataset}\label{sec:dataset_details}

\subsection{Image-Caption Collection}
We describe the list of models used for collection in Table~\ref{tab:caption_models}. All models except for closed ones are downloaded from Hugging Face~\cite{refB}.

\input{tables/models}

\noindent\textbf{Captioner models.} We collect data from two sources and employ five text-to-image models. The first source is CC12M, which is designed for vision-and-language pre-training and provides broad domain coverage. The second source is the COCO 2017 dataset, where we use the validation split. For both datasets, we cluster images into 50 domains based on ResNet features and then sample 40 images from each cluster, resulting in a total of 2,000 images per dataset. 

For the Captioner models, we randomly select one of the following instructions:

\begin{tcolorbox}[colback=gray!10,colframe=black!50,title=Instruction given to captioner models
]

1. Describe this image in detail.

2. Describe this image in detail. Instead of describing the imaginary content, only describe the content one can determine confidently from the image.

3. Provide a detailed description of the image, but only include elements that are clearly visible and verifiable.

4. Describe this image in detail. Minimize aesthetic descriptions as much as possible.

5. Provide a detailed, factual description without using emotional language.

\end{tcolorbox}

\noindent\textbf{Text-to-image models.} We employ five text-to-image models. The first is stabilityai/stable-diffusion-3.5-medium, a diffusion-based generative model that we run locally via the Diffusers library on GPU hardware. The others are OpenAI's gpt-image-1 and gpt-image-1.5, and Google's Gemini 2.5 Flash Image and Gemini 3 Pro Image, which are accessed through their respective APIs with gpt-4o-mini acting as the controller for image generation. For all models, we use identical prompts. To encourage category diversity, we predefine 170 object categories and randomly select one to be included in each prompt. The selected category is then inserted into an instruction given to gpt-4o-mini, which produces a 3--4 sentence description following the specification below.

\begin{tcolorbox}[colback=gray!10,colframe=black!50,title=Instruction given to GPT-4o-mini for producing text-to-image prompts
]
I want to create prompts to generate image using text to image model.
The prompts need to satisfy the following criteria.

1. The prompts include 3-4 sentences.

2. They need to describe a scene including {target}.

3. They need to describe the state of the objects, what they are doing.

4. They need to describe the location of the object in image, (e.g., left, right, bottom, top, etc)

5. They also need to describe where the objects are looking at (e.g., left, right, bottom, top, or towards some) if the object is some organism.

Can you suggest a prompt?
Please return in the form of dictionary, with a key of ``prompt''.

Output:
\end{tcolorbox}

\subsection{Voting and Quality Control}
We first recruited five annotators and conducted a pilot on one hundred images. The authors reviewed all annotations, and annotators who failed to meet our quality standards were not assigned further items. This process allowed us to identify trusted annotators.  
Each trusted annotator was then assigned between one thousand and two thousand images. The authors checked the quality for every batch of about two hundred images. If the annotations did not meet our standards, annotators were required to re-annotate before proceeding.

After the main annotation, we applied multi-round voting. Annotator-specific weights were assigned, with trusted annotators given higher weights. The aggregated votes were used to determine the final labels.  
For the \textit{incorrect} (hallucination) category, we adopted a stricter rule: if one trusted annotator or two annotators labeled an item as incorrect, the authors manually reviewed it, since hallucinations are more difficult to detect reliably than correctness.  
Finally, the authors adjudicated all ambiguous cases. This combination of pilot screening, ongoing audits, weighted voting, and final review ensured high-quality hallucination detection annotations.

For the 
Figure~\ref{fig:annotator_ui_detection} shows the annotation interface for the hallucination detection phase, while Figure~\ref{fig:annotator_ui_typing} shows the interface used for the hallucination type annotation phase.

\subsection{Annotation Quality}
We assess the quality of the annotations in \benchmarkname through manual inspection. Specifically, we randomly sample 100 images from different models, resulting in a total of 539 annotated sentences for evaluation. Each sentence is carefully examined by authors to verify whether it accurately describes the visual content of the corresponding image.
Our analysis shows that 97\% of the annotations are correct, indicating that the vast majority of descriptions accurately reflect the image content. These results confirm the reliability of our annotation pipeline.

\input{figures/annotator_ui_detection}
\input{figures/annotator_ui_typing}

\input{tables/hal_type}

\subsection{Hallucination Type and Location Annotation}\label{sec:annotation_type}
\input{figures/type_ex}

Table~\ref{tab:hallucination_types} shows the eight hallucination type categories used in the \benchmarkname.
These categories cover both fine-grained object- and attribute-level mistakes as well as broader contextual errors.  
Figure~\ref{fig:error_type_ex} shows annotation examples for each error type. Hallucinations are highlighted in red.

\subsection{Basic Statistics by Model}
\input{tables/i2t_stats}
\input{tables/t2i_stats}
Tables \ref{tab:dataset_stats_i2t} and \ref{tab:dataset_stats_t2i} show the basic statistics for each model. Most of the models have high positive-sentence rate. This indicates that, to obtain a sufficient number of hallucination examples from real captioner outputs, a large number of sentences is required. Our dataset contains multiple sentences per image and detailed sentence descriptions, making it suitable for more fine-grained analysis.

\input{tables/comparison}

\subsection{Additional Analysis}
\noindent\textbf{Detailed comparison against existing datasets.} 
Table~\ref{tab:dataset_compare} describes the detailed comparison against prior hallucination detection datasets applicable for HalDec. Our dataset includes more responses and includes text-to-image models as the evaluation target. 
In particular, it offers larger textual coverage, covering 104k sentences, and a vocabulary of 17.5k unique word types, than prior datasets.

\input{figures/hal_image_domain}
\noindent\textbf{Image domain.} 
Figure~\ref{fig:hul_by_type_CC12M} illustrates the ratio of incorrect sentences on each image category. Most Captioners tend to produce more errors in \textit{Text} and \textit{Illustration} domains, while they are relatively robust in real images like \textit{Buildings} and \textit{Scenary}. This can be because of the bias in the training data of the Captioners.

\input{figures/incorrect_position}

\noindent\textbf{Error analysis w.r.t position of the sentence.} 
In Fig.~\ref{fig:incorrect_position}, we present the ratio of incorrect sentences across sentence positions for each model. 
Among image captioning models, incorrect sentences tend to appear most frequently in the second to fourth positions. Interestingly, the very first sentence is less likely to contain hallucinations. This may be because the first sentence often serves as an overall image caption. In contrast, the second and subsequent sentences typically provide more detailed descriptions, which are more prone to errors. 
For positions beyond the sixth sentence, the error rate decreases again. 
These later sentences often serve as overall conclusions or closing remarks rather than detailed descriptions, which may make them similar to the first sentence and thus less prone to errors.

\input{figures/error_type}
\noindent\textbf{Analysis w.r.t hallucination types.} 
Figure~\ref{fig:error_type} describes the type of hallucinations we provide. Our dataset covers various kinds of hallucinations. 

\section{Details of Experimental Setups}
\subsection{Details of Evaluation}\label{sec:details_eval}
\noindent\textbf{Source of models.} We employ models available in HuggingFace and base our code on the HuggingFace Transformers package. 

\noindent\textbf{Computation.} At most eight A100 80GB GPUs are used for inference of a single model. 

\noindent\textbf{Prompt.} We employ the prompt below to compute the alignment score for decoder-based VLM. 
\begin{tcolorbox}[colback=gray!10,colframe=black!50,title=Prompt to compute image-sentence alignment]
You are given an image and a caption describing the given image. 
Your task is to judge if the caption describes the image correctly. 
If you think the sentence does not describe the image correctly, return low the score. 
If you think there is no mistake in the caption, return high score. Judge the correctness from 0-100 points. 
Return the output in the form of dictionary, e.g., {``score'': 50}. Please first output the correctness points before explaining the reason for the score.  

Caption: 
\end{tcolorbox}

Similarly, we use the prompt below to obtain the results of the chain of thought. 
\begin{tcolorbox}[colback=gray!10,colframe=black!50,title=Chain-of-thought prompt]
You are given an image and a caption describing the given image. 
Your task is to judge if the caption describes the image correctly. If you think the sentence does not describe the image correctly, return low the score. 
If you think there is no mistake in the caption, return high score.
Judge the correctness from 0-100 points. 
Return the output in the form of dictionary, e.g., {``score'': 50}.
Please first explain the reason of scoring in ** two or three **  sentences and output the correctness points as shown above. 

Caption: 
\end{tcolorbox}

\noindent\textbf{Parsing.} After obtaining the text output, we write a parser to convert the output into an integer. Models sometimes did not properly follow the prompt, and we could not parse such output. For such a sample, we assign 50 as its alignment score. In Table~\ref{tab:auc_compare}, we present models with their failure rate less than 5\%. Also, the failure rate of a well-performing model is very low. 

\noindent\textbf{Annotation details in self-preference analysis.} In Sec.~\ref{sec:detailed_analysis}, we additionally provide sentence-level hallucination existence labels for Qwen-2.5 (32B) and Gemma-3 (27B). To reduce the cost of annotation, we follow an annotation procedure different from the other 8 models, yet in a quality-ensured manner. Specifically, we randomly pick 500 images and generate captions using two models. Then, one quality-ensured annotator gives an annotation to 500 captions. This produces enough samples for analysis. We will include this split when publishing the dataset.

\noindent\textbf{Prompt in hallucination localization.} 
We employ the prompt below to obtain the results of hallucination localization.

\begin{tcolorbox}[colback=gray!10,colframe=black!50,title=Prompt for hallucination localization]
You are given an image and a caption describing the given image. 
Your task is to localize the segment of the caption, which describes the image incorrectly. 
Please output the segment by marking the incorrect parts by **[]**, e.g., A **[red]** bird singing in a tree.
Return the output in the form of a dictionary. 
Example format. 
\begin{verbatim}
```json
{
"output": "A **[red]** bird singing in a tree."
}
```
\end{verbatim}
Caption: 
\end{tcolorbox}

\noindent\textbf{Evaluation metric in hallucination localization.} We evaluate the alignment between the word spans predicted by models and the ground-truth (GT) spans using an Intersection-over-Union (IoU) based criterion. Concretely, we compute the IoU between the predicted word range and the GT word range. In Table~\ref{tab:localization_combined}, a prediction is considered correct if its IoU with a GT span is greater than or equal to 0.3. Based on this criterion, we measure precision as the proportion of predicted spans that are judged correct.

\section{Additional Experiments}
\noindent\textbf{Effectiveness of VLM-based filtering.}
\input{tables/filtering_exp}
Table~\ref{tab:filtering_exp} shows the effectiveness of caption filtering. We compare CLIP models fine-tuned on image–caption pairs that are either randomly selected or filtered using Qwen-3-VL. Specifically, we first generate 5M captions using Qwen-2 and then select approximately 1M captions based on the image–text alignment scores predicted by Qwen-3-VL. The result indicates that applying filtering can improve the performance of the model in most cases.

\input{tables/context_comparison}
\noindent\textbf{Does incorporating the preceding sentences lead to improved performance?} In our main paper, we focus on the evaluation of alignment between a single sentence and an image. This setting ignores the context from the preceding sentence in a caption since the model does not see the preceding sentences during inference. We study the effectiveness of adding the preceding sentences. Specifically, we feed all preceding sentences as well as the target sentence for evaluation as a prompt, \eg, \textit{a cat is running in a park. The cat is next to a kid. Think about the correctness of **The cat is next to a kid. **}. Table \ref{tab:context_compare} compares the effect of adding contextual sentences. Surprisingly, we find that providing this additional context does not clearly improve image–text alignment. For Llama-4, adding context often degrades instruction following, leading to more frequent parsing errors. For GPT5-mini, no parsing errors are observed, but the performance still drops slightly. These results suggest that incorporating preceding sentences as context can interfere with judging the alignment between the target sentence and the image. Since not all language models excel at handling long sequences, evaluation on a single sentence can be fair in evaluating the image-text alignment ability. 

\input{tables/cot}
\noindent\textbf{Chain-of-Thought improves the performance?} 
Table~\ref{tab:cot} evaluates the impact of chain-of-thought reasoning
~\cite{refD}, where detectors are prompted to generate a reasoning path before producing a score (see above for prompt details). For Llama-4, COT generally improves performance, whereas for some Captioners, the gains are marginal or even slightly negative. Results for GPT4.1-mini are mixed, wherein improvements highly depend on the evaluation target. 
\input{tables/self_ensemble}

\noindent\textbf{Self-ensemble improves performance.} 
We further study the potential of ensembling. Unlike the analysis above, we ensemble outputs from a single model to refine detector's score~\cite{refE,refF}.
To get different scores from a single model, we obtain different reasoning paths by stochastic sampling in the chain-of-thought. To ensure the diversity of COT, we set the temperature as 1.5 and top$_\text{p}$ as 0.9. 
Table~\ref{tab:self_ensemble} presents the results in Llama-4, where the performance consistently improves in all Captioners. Also, using more ensemble paths tends to improve the performance, while the increase seems to saturate. Model ensembling can be an interesting direction to improve the performance in this task. 
\input{tables/compare_existing_methods}

\noindent\textbf{VLM detectors can surpass prior approaches.} Table~\ref{tab:compare_other_method} presents the comparison to UniHD~\cite{chen2024unified}, which prompts LLM to utilize an open-vocabulary detector and OCR engine. The results indicate that advanced VLMs can surpass the approach without using such external tools. More detailed discussion is available in the appendix.

\input{tables/localization_miou}

\noindent\textbf{Mean intersection over union in hallucination localization.} Table~\ref{tab:mean_iou} shows the results of mean IoU in hallucinated segment localization. Specifically, we compute the intersection over union between the predicted and ground-truth segments and compute the average for all samples. Overall, the performance is consistent with what is reported in Table~\ref{tab:localization_combined}.

\input{figures/self_pref_score_appendix}

\noindent\textbf{Additional results in self-preference evaluation.} 
Figure~\ref{fig:self_pref_bias_appendix} illustrates self-preference score analysis for Gemma-27B, Llama-4, and Qwen2.5. Their self-preference tendency is significant, especially for Gemma-27B and Qwen2.5. 


\input{figures/score_distribution}
\noindent\textbf{Distributions of evaluators' outputs.}
Figure~\ref{fig:score_dist} illustrates the distribution of evaluators' scores for GPT-4o captions. Scores tend to concentrate on the points near 0 and 100 for Llama-4.

\input{figures/ex_scoring_appendix}

\noindent\textbf{Additional examples of VLMs' outputs.} Figure~\ref{fig:ex_scoring_appendix} illustrates examples of input images, sentences, and corresponding correctness scores inferred by VLMs. VLMs tend to make errors in the location of the objects, the relationship between them, and small visual details.

\section{Additional Examples of Annotations}\label{sec:annotation_example_appendix}
We provide additional figures illustrating annotation results and representative hallucination cases: ShareGPT (Fig.~\ref{fig:error_type_ex_sharegpt}), LLaVA (Fig.~\ref{fig:error_type_ex_llava}), Qwen-2 (Fig.~\ref{fig:error_type_ex_gpt4o}), GPT-4o (Fig.~\ref{fig:error_type_ex_gpt4o}), CogVLM (Fig.~\ref{fig:error_type_ex_cogvlm}), LLaMA-4 (Fig.~\ref{fig:error_type_ex_llama4}), Stable Diffusion (Fig.~\ref{fig:error_type_ex_sd}), and GPT-Gen (Fig.~\ref{fig:error_type_ex_gptgen}).

\input{figures/ann_ex_sharegpt}
\input{figures/ann_ex_llava}
\input{figures/ann_ex_qwen}
\input{figures/ann_ex_gpt4o}
\input{figures/ann_ex_cogvlm}
\input{figures/ann_ex_llama-4}
\input{figures/ann_ex_sd}
\input{figures/ann_ex_gpt_gen}



%
%

\makeatletter
\let\BR@@lbibitem\@lbibitem
\let\BR@@bibitem\@bibitem
\makeatother

%% file: tables/models.tex
\begin{table}[h]
  \small
    \caption{
      \small
Details of VLMs picked as Captioners and Text2Image models. We cover diverse models considering their size, provider, and release date.}
  \centering
  \resizebox{0.8\textwidth}{!}{%
  \begin{tabular}{lcccc}
  \toprule
\textbf{Model} & \textbf{Provider} & \textbf{Open/Closed} & \textbf{Scale} & \textbf{Release} \\
\midrule
GPT-4o & OpenAI & Closed & - & 2024/05 \\
ShareGPT (Share Captioner) & Shanghai AI Laboratory & Open & 7B & 2023/11 \\
LLaVA-NeXT (llava-next-72b-hf) & Microsoft & Open & 72B & 2024/01 \\
Llama-4-Scout (17B-16E) & Meta & Open & 109B & 2025/04 \\
Qwen2.5-VL (7B-Instruct) & Alibaba & Open & 7B & 2024/12 \\
CogVLM (
cogvlm2-llama3-chat-19B ) & Tsinghua Univ. & Open & 19B& 2024/06 \\
GPT-5&OpenAI & Closed & - & 2025/08\\
Gemini-3-Pro& Google & Closed & - & 2025/11\\
Qwen 2.5& Alibaba & Open & 32B & 2024/09\\
Gemma-3&Google&Open&27B&2025/03\\
Stable-diffusion-3.5-medium (SD) & Stability AI & Open & 2.5B &2024/10\\
GPT-Image-1 (GPT4o-mini) & OpenAI & Closed & - & 2024/05\\
GPT-Image-1.5~(GPT-Img-1.5)& OpenAI & Closed & - &2025/12\\
Gemini 2.5 Flash Image~(Gemini-2.5)& Google & Closed & - & 2025/08\\
Gemini 3 Pro Image~(Gemini-3)& Google & Closed & - & 2025/11\\
    \bottomrule
    \hline
    \end{tabular}%
    }
    
  \label{tab:caption_models}%
\end{table}%

%% file: figures/annotator_ui_detection.tex
\begin{figure*}[t]
\centering
\includegraphics[width=\linewidth]{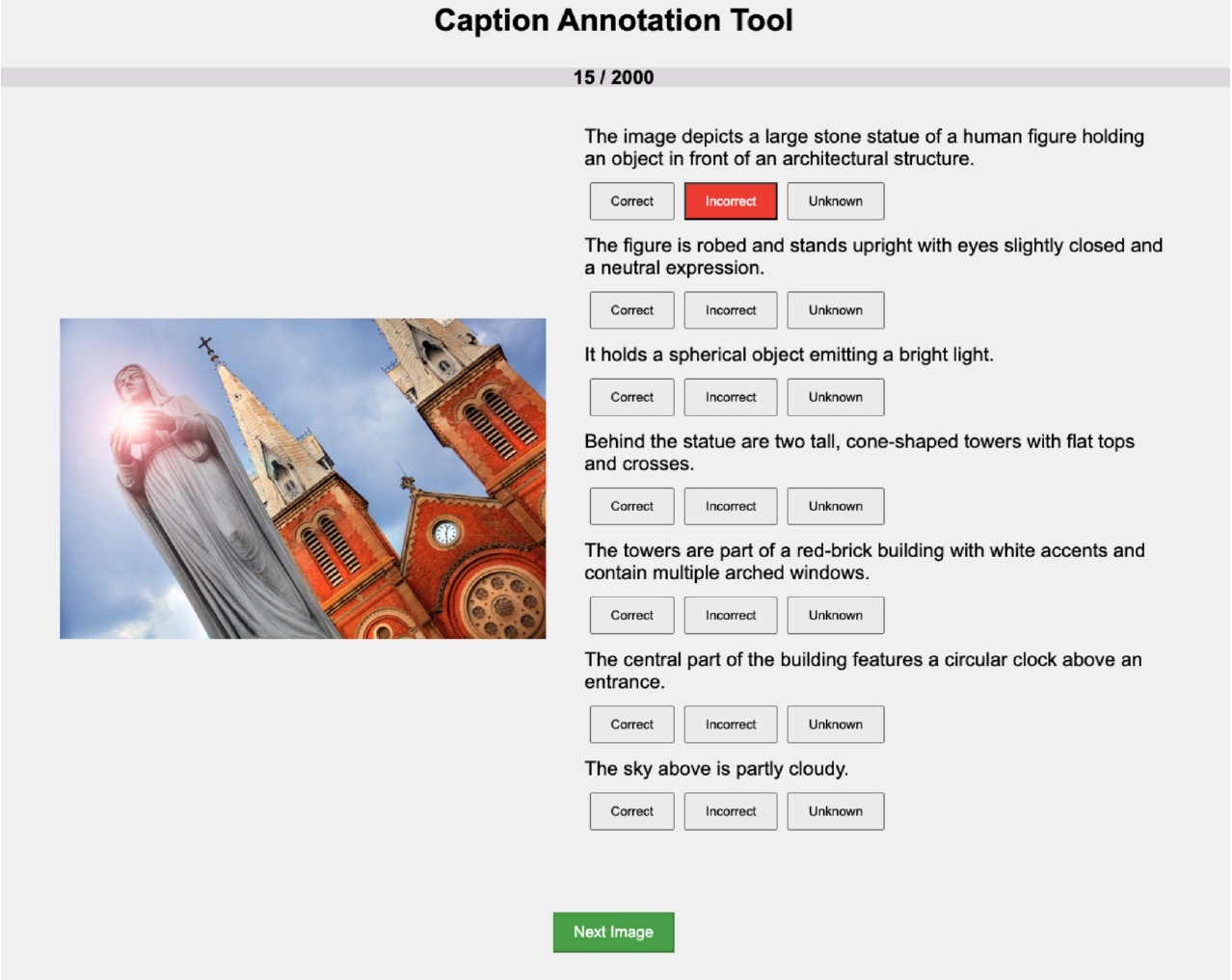}
\caption{Example of an interface used for the hallucination detection.}
\label{fig:annotator_ui_detection}
\end{figure*}

%% file: figures/annotator_ui_typing.tex
\begin{figure*}[t]
\centering
\includegraphics[width=\linewidth]{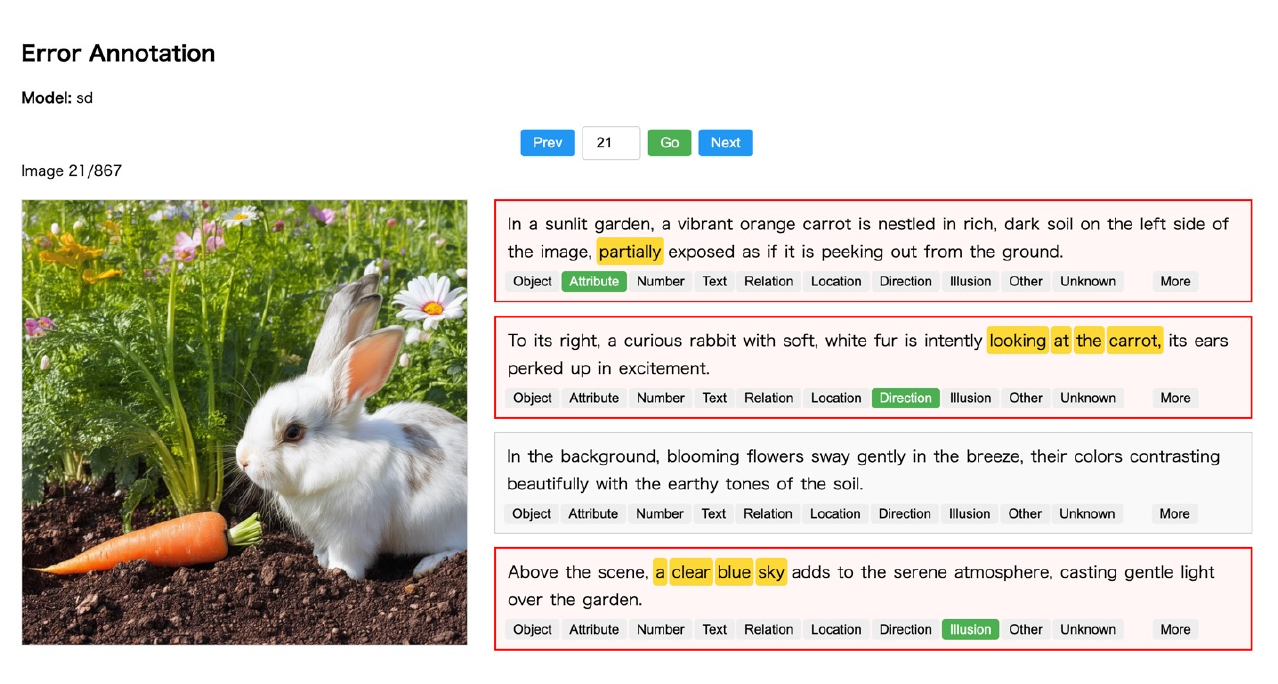}
\caption{Example of an interface used for the hallucination type annotation.}
\label{fig:annotator_ui_typing}
\end{figure*}

%% file: tables/hal_type.tex
\begin{table*}[t]
  \small
   \caption{
    \small
    Types of hallucinations categorized for analysis.
  }
  \centering
  \resizebox{\textwidth}{!}{%
    \begin{tabular}{ll}
        \toprule
        \textbf{Type} & \textbf{Description} \\
        \midrule
        Object & Misidentifies an object or uses an incorrect noun (e.g., calling a dog a cat). \\
        Attribute & Incorrect description of an object's property such as color, size, or action (e.g., red car described as blue). \\
        Number & Incorrectly states the number of objects or people (e.g., “three people” when only two are present). \\
        Text & Misreads or misrepresents textual information in the image (e.g., misreading a store sign). \\
        Relation & Incorrect description of relationships between objects (e.g., “a man riding a horse” when he is standing next to it). \\
        Location & Misrepresents the position of an object in the image (e.g., “a cup on the table” when it is on the floor). \\
        Direction & Incorrectly describes the direction/orientation of an object (e.g., “a person facing left” when they face right). \\
        Illusion & Describes objects, scenes, or actions that do not exist at all (e.g., mentioning “a flying bird” when no bird is present). \\
        \bottomrule
    \end{tabular}%
  }

  \label{tab:hallucination_types}
\end{table*}

%% file: figures/type_ex.tex
\begin{figure*}[t]
\centering
\includegraphics[width=\linewidth]{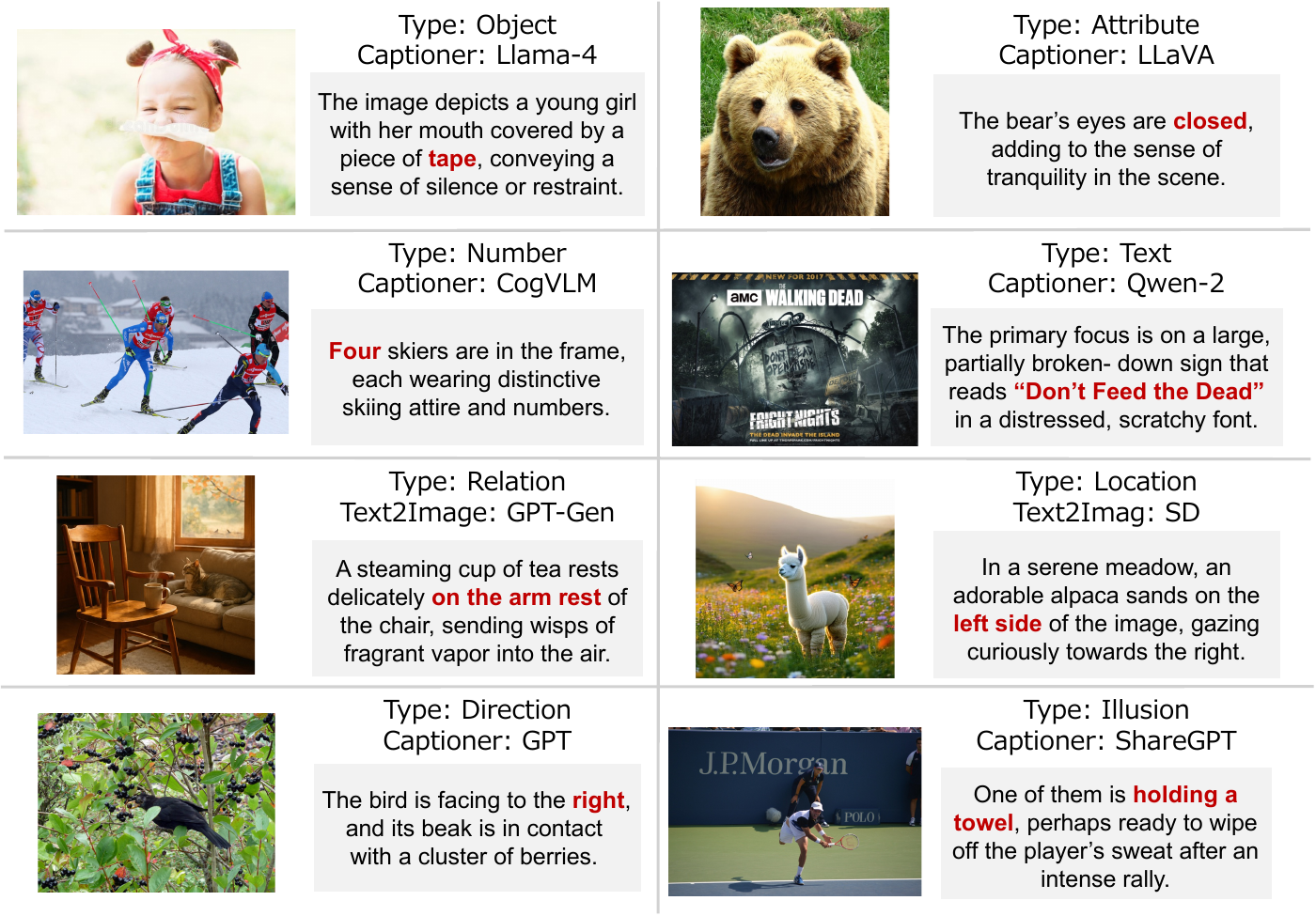}
\caption{Example annotations of error type. Hallucinations are highlighted in red.}
\label{fig:error_type_ex}
\end{figure*}

%% file: tables/i2t_stats.tex
\begin{table*}[t]
   \caption{
      \small \textbf{Stats of \benchmarkname in sentence-level correctness annotations on image to text models.} 
      We exclude sentences with \textit{unknown} label.}
      \vspace{-3mm}
  \centering
    \fontsize{7.2pt}{8.0pt}\selectfont
    \setlength\tabcolsep{5.0pt} 
    \renewcommand{\arraystretch}{1.5} 
  \resizebox{0.95\textwidth}{!}{%
    \begin{tabular}{lccccccccccc}
        \toprule
      & CogVLM & GPT4o &ShareGPT (S-GPT)  & Llama-4 & LLaVA-NeXT & Qwen-2 &Gemini-3-Pro&GPT-5&Gemma-3&Qwen 2.5 & Total\\ 
      \midrule    
      Sentences & 7372 & 12790 & 17610 & 14800 & 15170  & 15790 & 2483&2228&2337&2600&93180\\
      Sentences / Image & 3.7 &  6.4 & 8.8&7.4&6.9&7.9&5.0&4.5&4.7&5.2&6.7\\
      Positive / Total (\%) & 91.5 & 91.8 & 73.4 & 85.2 & 80.9 & 88.8 &92.4& 93.5&90.1&92.2&85.0 \\
      Word / Sentence & 15.9 & 14.7 & 18.3 & 19.1 & 17.4 & 17.3 &17.4&15.5&16.6&18.4 &17.2\\
            \bottomrule   
    \end{tabular}%
  }       
  \label{tab:dataset_stats_i2t}
\end{table*}

%% file: tables/t2i_stats.tex
\begin{table*}[t]
   \caption{
      \small \textbf{Stats of \benchmarkname in sentence-level correctness annotations on text to image models.} \benchmarkname contains a large number of annotated sentences, enough for benchmarking models. We exclude sentences with \textit{unknown} label.}
      \vspace{-3mm}
  \centering
    \fontsize{7.2pt}{8.0pt}\selectfont
    \setlength\tabcolsep{5.0pt} 
    \renewcommand{\arraystretch}{1.5} 
  \resizebox{0.95\textwidth}{!}{%
    \begin{tabular}{lcccccc}
        \toprule
      & Stable-diffusion (SD) & GPT-Image-1 (GPT4o-mini) & GPT-Image-1.5 (GPT-Img-1.5) & Gemini 2.5 Flash Image (Gemini-2.5) & Gemini 3 Pro Image (Gemini-3) &All\\ 
      \midrule    
      Sentences & 2415&3524&1764&1761&1693&11157\\
      Sentences / Image & 2.4&3.5&3.5&3.5&3.4&3.2\\
      Positive / Total (\%) & 34.1&79.7&90.6&87.8&91.0&74.5 \\
      Word / Sentence &  22.9&22.5& 22.4 &22.4&22.4&22.6\\
            \bottomrule   
    \end{tabular}%
  }       
  \label{tab:dataset_stats_t2i}
\end{table*}

%% file: tables/comparison.tex

\begin{table*}[h]
  \small
  \caption{
      \small
     Compared to existing hallucination detector benchmarks for image captions based on their evaluation split, \benchmarkname offers large number of responses with actual Captioner's hallucination annotated by human.
     This scale enables detailed, model-wise performance analysis and facilitates a deeper understanding of detector characteristics. For datasets that are not publicly available or lack information, the corresponding statistics are reported as NA.}
    \centering    
   \scalebox{0.72}{
    \begin{tabular}{lcccccccccccccc}
    \hline
    \toprule
    \textbf{Dataset} & \makecell{Segment\\Annotation} &
    \makecell{\# halluc.\\types} &
    \makecell{\# models} &
    \makecell{\# sentences} &
    \makecell{ words/Sent.} &
    \makecell{\# vocab.} &
    \makecell{\# unique\\image} &
    \makecell{Sent. Generation} \\
    \midrule
    Foil~\cite{petryketal2024aloha}        &\checkmark   & \colorxmark&0&5k & 11.8 & 4.1k &2.5k& Rule-base \\
HAT~\cite{petryketal2024aloha}         &    \checkmark     & \colorxmark&1&0.4k & 13.6 & 1.2k  &0.4k& Captioner \\
ARO~\cite{yuksekgonul2022and} &  \checkmark &\checkmark &0&52k &7.6  & 1.5k  &6.6k& Rule-base \\
Winoground~\cite{thrush2022winoground}   &  \checkmark    &3&0&1.6k& 9.0  & 0.9k  &0.8k& Human\\
SugarCrepe~\cite{hsieh2023sugarcrepe}  & \colorxmark&3  &0&2k  & 11.1 & 2.2k &1.5k& Language Model \\
SeeTrue~\cite{yarom2023you}     &   \checkmark   &\colorxmark&1&6.9k& 11.5 & 1.5k &6.9k&T2I model\\
GenAI-Bench~\cite{lin2024evaluating} & \colorxmark   &6&10&1.6k & 12.6 & 4.3k &9.6k&T2I model\\
MHalDetect~\cite{gunjal2024detecting}  & \checkmark  & \colorxmark&1&14k&18.0 & 4.4k &0.8k&Captioner\\

    HaELM~\cite{wang2023evaluation} & \colorxmark &  \colorxmark & 3&1.5k&13.7& 1.6k&NA& Captioner\\
    MHaluBench~\cite{chen2024unified} & \checkmark  &  4 & 8 &0.7k&14.6&1.3k&0.2k& Captioner + T2I\\  
    HalLoc~\cite{refG}&  \checkmark &4&0&155.9k&17.7&NA&NA&GPT-4 Injection\\  
    ZINA~\cite{wada2025zina} & \checkmark  &  6 & 12 &NA&NA&NA&NA&Captioner & \\
    \midrule
    \benchmarkname (Ours) & \checkmark  &9&15&104k&17.8&17.5k&5.5k& Captioner + T2I\\  
    \bottomrule
    \hline
    \end{tabular}
    }
      
  \label{tab:dataset_compare}
\end{table*}


%% file: figures/hal_image_domain.tex
\begin{figure*}[t]
\centering
\includegraphics[width=0.95\linewidth]{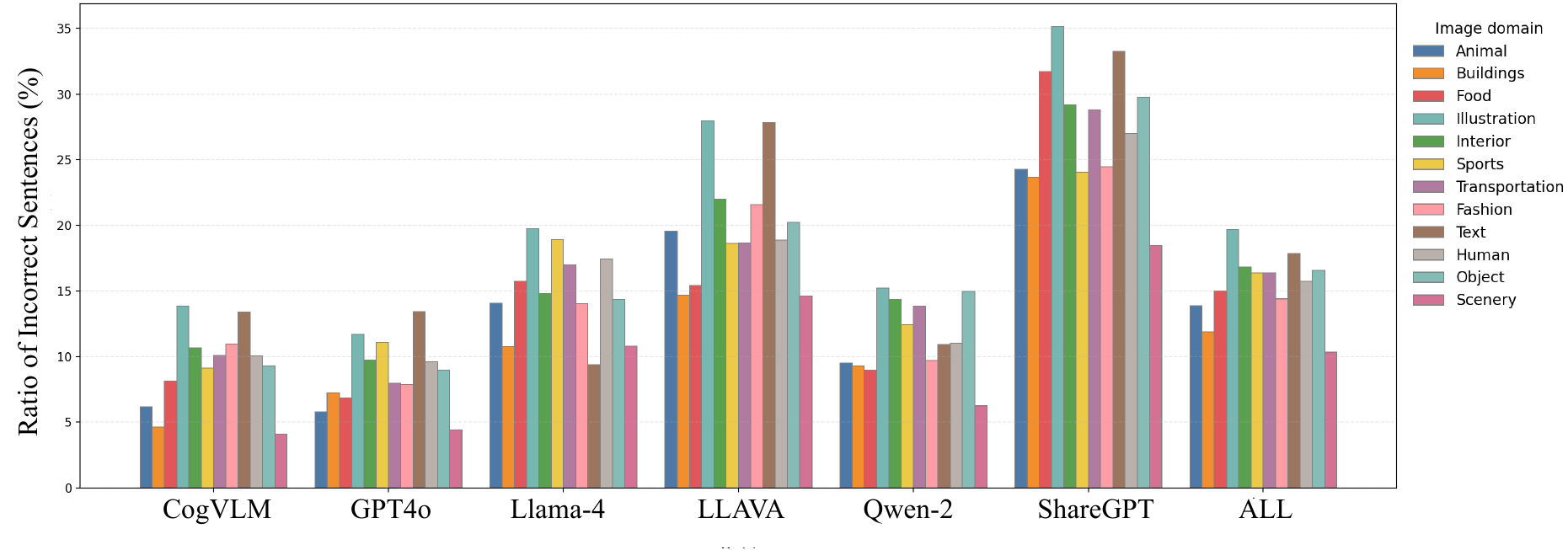}
\caption{\small \textbf{Ratio of incorrect sentences for each image domain.} All models tend to produce more errors in domains such as \textit{illustration} and \textit{Text}.}
\label{fig:hul_by_type_CC12M}
\end{figure*}

%% file: figures/incorrect_position.tex
\begin{figure*}[t]
\centering
\includegraphics[width=0.95\linewidth]{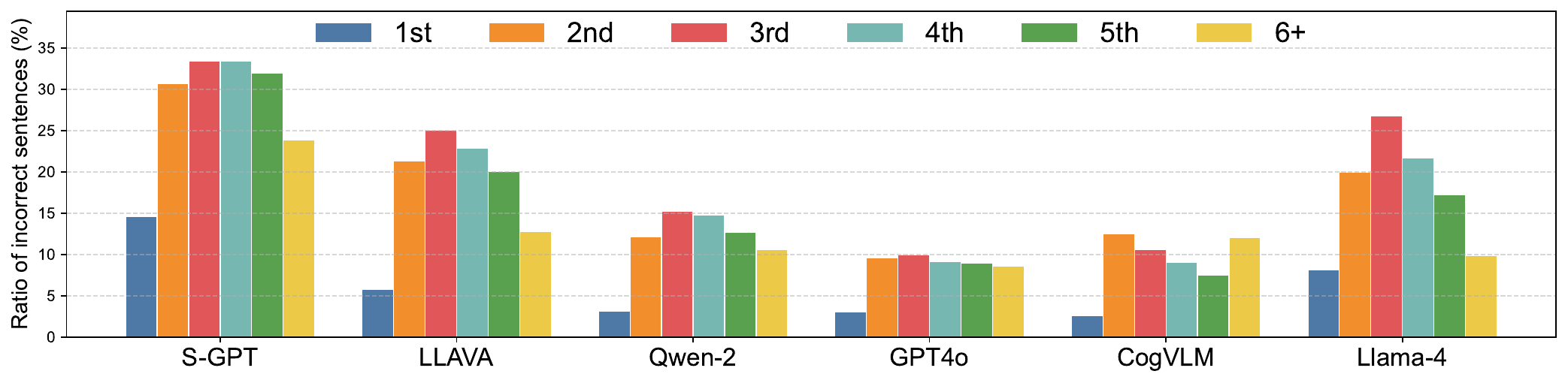}
\vspace{-4mm}
\caption{\small \textbf{Ratio of incorrect sentences within each sentence position per model.} Different colors indicate different positions. All models produce fewer errors at the 1st position.}
\label{fig:incorrect_position}
\end{figure*}

%% file: figures/error_type.tex
\begin{figure*}[t]
\centering
\includegraphics[width=0.8\linewidth]{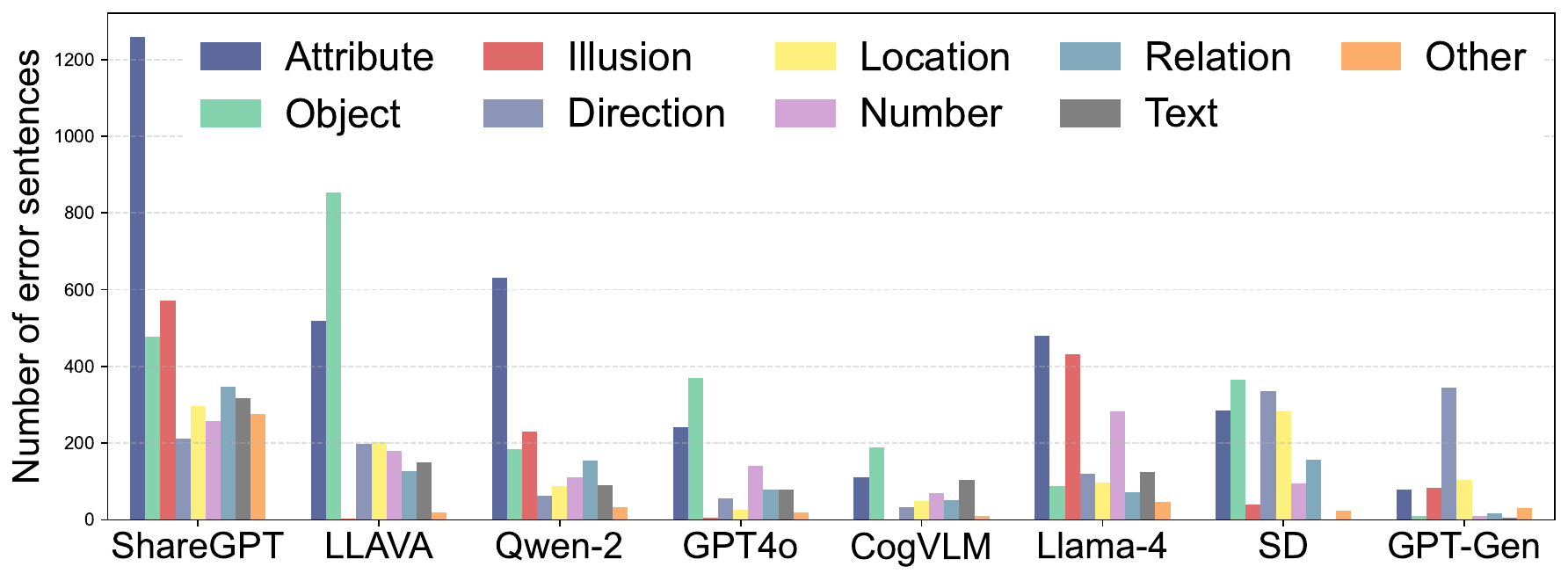}
\vspace{-3mm}
\caption{\small \textbf{Number of hallucinations for each category.} Most models make many mistakes in attributes and text.}
\label{fig:error_type}
\end{figure*}

%% file: tables/filtering_exp.tex
\begin{table}[t]
\centering
\small
\caption{Image-text retrieval performance (Recall@5). We fine-tune CLIP using captions generated by Qwen-2. By applying filtering based on Qwen-3, we see improvements on COCO and Flickr8k retrieval. }
\vspace{-2mm}
\fontsize{6.2pt}{9.0pt}\selectfont
\setlength\tabcolsep{5.0pt}
\renewcommand{\arraystretch}{0.95}

\begin{tabular}{l|cc|cc}
\toprule
\multirow{2}{*}{Method} 
& \multicolumn{2}{c|}{COCO} 
& \multicolumn{2}{c}{Flickr8k} \\
\cmidrule(lr){2-3}\cmidrule(lr){4-5}
& T2I & I2T & T2I & I2T \\
\midrule
No tuning & 54.5 & 74.0 & 80.4 & 90.2 \\
Random Selection & 56.3 & 75.8 & 81.7 & \textbf{93.6} \\
Qwen3 Filtering & \textbf{58.6} & \textbf{77.1} & \textbf{84.1} & 93.0 \\
\bottomrule
\end{tabular}
\label{tab:filtering_exp}
\end{table}

%% file: tables/context_comparison.tex
\begin{table*}[h]
  \centering
   \caption{\small Results of using preceding sentences as context.} 
    \fontsize{7.2pt}{8.0pt}\selectfont
    \setlength\tabcolsep{5.0pt} 
    \renewcommand{\arraystretch}{1.5} 
  \scalebox{0.9}{
    \begin{tabular}{c|c|ccc}
    \toprule
    \textbf{Detector}& \textbf{Context}&  \textbf{S-GPT} & \textbf{Llava} &
    \textbf{CogVLM} \\\hline            
   \multirow{2}{*}{Llama-4 (109B)}  && \textbf{80.7} & 78.6 &\textbf{77.2}\\
   &\checkmark & 80.6& \textbf{79.1} & 73.5\\\arrayrulecolor{gray!40}\cline{1-4}\arrayrulecolor{black}  
    \multirow{2}{*}{GPT5-mini} &  &81.5&\textbf{82.2}&\textbf{81.1}\\
    &\checkmark &\textbf{82.6}&81.5&79.5\\
    \bottomrule
    \end{tabular}
  } 
  \label{tab:context_compare}
\end{table*}

%% file: tables/cot.tex
\begin{table*}[h]
  \centering
   \caption{\small Results of using Chain-of-Thought (COT).} 
    \fontsize{7.2pt}{8.0pt}\selectfont
    \setlength\tabcolsep{5.0pt} 
    \renewcommand{\arraystretch}{1.5} 
  \scalebox{0.67}{
    \begin{tabular}{c|c|llllll|ll}
    \toprule
   \multirow[c]{2}{*}{{\textbf{Detector}}} &
    \multirow[c]{2}{*}{{\textbf{COT}}} &   
    \multicolumn{6}{c|}{\textbf{Image-to-Caption Models}} & 
    \multicolumn{2}{c}{\textbf{Text-to-Image Models}} \\
    \cmidrule(lr){3-8} \cmidrule(lr){9-10}
     & &  \textbf{S-GPT} & \textbf{Llava} & \textbf{Qwen-2} & \textbf{GPT4o} &
    \textbf{CogVLM} & \textbf{Llama-4} & \textbf{SD} & \textbf{GPT-Gen} \\\hline            
   \multirow{2}{*}{Llama-4 (109B)}  &  &  80.7 & 78.6& 77.6& 67.5& 77.2& 59.9 &	81.1 & 64.7\\
   &\checkmark & 80.6\minusvalue{-0.1} & 80.8\plusvalue{2.2} & 80.0\plusvalue{2.4} & 71.1\plusvalue{3.6}  & 80.1\plusvalue{2.9}&62.4\plusvalue{2.5}	&80.8\minusvalue{-0.3} & 65.1\plusvalue{0.4} \\\arrayrulecolor{gray!40}\cline{1-10}\arrayrulecolor{black}  
    \multirow{2}{*}{GPT4.1-mini} &  & 77.8 & 75.8 & 74.4 & 65.8 & 69.2 & 66.0 & 68.7 & 56.1\\
     & \checkmark &  79.0\plusvalue{1.2} & 76.2 \plusvalue{0.4} & 75.0\plusvalue{0.6}& 63.4\minusvalue{-2.4} & 71.6\plusvalue{2.4} & 63.8\minusvalue{-2.2}& 73.2\plusvalue{4.5}& 56.2\plusvalue{0.1}\\
    \bottomrule
    \end{tabular}
  } 
  \label{tab:cot}
\end{table*}

%% file: tables/self_ensemble.tex
\begin{table*}[h]
  \centering
   \caption{\small Ensembling detectors' output improves performance in almost all cases. We highlight the increase or decrease from the \textit{better} model used for ensembling next to each score.} 
    \fontsize{7.2pt}{8.0pt}\selectfont
    \setlength\tabcolsep{5.0pt} 
    \renewcommand{\arraystretch}{1.5} 
  \scalebox{0.62}{
    \begin{tabular}{cc|llllll|ll}
    \toprule
   \multirow[c]{2}{*}{{\textbf{Detector}}} &
    \multirow[c]{2}{*}{{\textbf{Num. of Ensemble}}} &   
    \multicolumn{6}{c|}{\textbf{Image-to-Caption Models}} & 
    \multicolumn{2}{c}{\textbf{Text-to-Image Models}} \\
    \cmidrule(lr){3-8} \cmidrule(lr){9-10}
     & &  \textbf{S-GPT} & \textbf{Llava} & \textbf{Qwen-2} & \textbf{GPT4o} &
    \textbf{CogVLM} & \textbf{Llama-4} & \textbf{SD} & \textbf{GPT-Gen} \\\hline            
   Llama-4-scout & 1 & 80.6 & 80.8 & 80.0 & 71.1&	80.1 & 62.4 & 80.8 & 65.1 \\
   Llama-4-scout & 5 &  83.2\plusvalue{2.6} & 83.0\plusvalue{2.2}& 81.8 \plusvalue{1.9}& 74.4\plusvalue{3.4}&82.2\plusvalue{2.1} &  65.1\plusvalue{2.7}& 83.0 \plusvalue{2.2}& 65.9\plusvalue{0.8} \\
   Llama-4-scout & 10 &  83.7\plusvalue{3.1} & 83.4\plusvalue{2.6} & 82.2 \plusvalue{2.3} & 75.0\plusvalue{3.9} & 82.8\plusvalue{2.7} &65.7\plusvalue{3.3} & 83.4 \plusvalue{2.6} & 66.4\plusvalue{1.3}   \\
    \bottomrule
    \end{tabular}
  }
 
  \label{tab:self_ensemble}
\end{table*}

%% file: tables/compare_existing_methods.tex
\begin{table*}[t]
  \small
   \caption{\small Comparison to existing HalDec approaches. }
  \centering
    \fontsize{7.2pt}{8.0pt}\selectfont
    \setlength\tabcolsep{5.0pt} 
    \renewcommand{\arraystretch}{1.5} 
  \scalebox{0.95}{
    \begin{tabular}{c|cc}
    \toprule
   \textbf{Detector} & \textbf{GPT4o} & \textbf{SD}  \\\hline
    UniHD~\cite{chen2024unified}&62.6&71.0\\\hline
    Qwen-2.5 32B & 66.1 & 68.9\\   
    Gemma-3 27B & 61.0 & 63.7\\
    Llama-4 109B & 67.7 & 81.1\\
    GPT-4.1-mini &65.8& 68.7\\
    \bottomrule
    \end{tabular}
  } 
  \label{tab:compare_other_method}
\end{table*}

%% file: tables/localization_miou.tex
\begin{table*}[t]
  \centering
   \caption{\small Mean IoU for hallucination localization task. Localizing the segment of the hallucinated caption remains difficult even for performant models.} 
    \fontsize{7.2pt}{8.0pt}\selectfont
    \setlength\tabcolsep{5.0pt} 
    \renewcommand{\arraystretch}{1.5} 
  \scalebox{0.7}{
    \begin{tabular}{ccccccccccccc}
    \toprule
   \multirow[c]{2}{*}{{\textbf{Detector}}} &   \multirow[c]{2}{*}{{\textbf{Params}}} & &
    \multicolumn{6}{c}{\textbf{Image-to-Caption Models}} & &
    \multicolumn{2}{c}{\textbf{Text-to-Image Models}} & \multirow[c]{2}{*}{{\textbf{Avg.}}} \\
    \cmidrule(lr){4-9} \cmidrule(lr){11-12}
     & & & \textbf{S-GPT} & \textbf{Llava} & \textbf{Qwen-2} & \textbf{GPT4o} &
    \textbf{CogVLM} & \textbf{Llama-4} && \textbf{SD} & \textbf{GPT-Gen} \\\hline    
    Qwen-2.5 & 32B & & 13.8 & 15.1 & 11.7 & 15.1 & 16.0 & 10.7 && 11.4 & 9.4 & 12.9 \\
     GPT-4o mini & - & & 21.6 & 22.4 & 18.3 & \textbf{23.3} & \textbf{21.5} & 16.4 && \textbf{12.5} & \textbf{11.9} & 18.5 \\
    Llama-4 & 109B & & 22.6 & 20.8 & 22.7 & 26.4 & 23.2 & 17.3 && 10.7 & 9.0 & 19.1 \\
    Llama-4 & 400B & & \textbf{24.8} & \textbf{22.1} & \textbf{23.3} & 26.0 & 21.7 & \textbf{18.0} && 11.9 & 9.3 & \textbf{19.6} \\   
    \bottomrule
    \end{tabular}
  }   
  \label{tab:mean_iou}
\end{table*}

%% file: figures/self_pref_score_appendix.tex
\begin{figure*}[h]
\centering
\includegraphics[width=\linewidth]{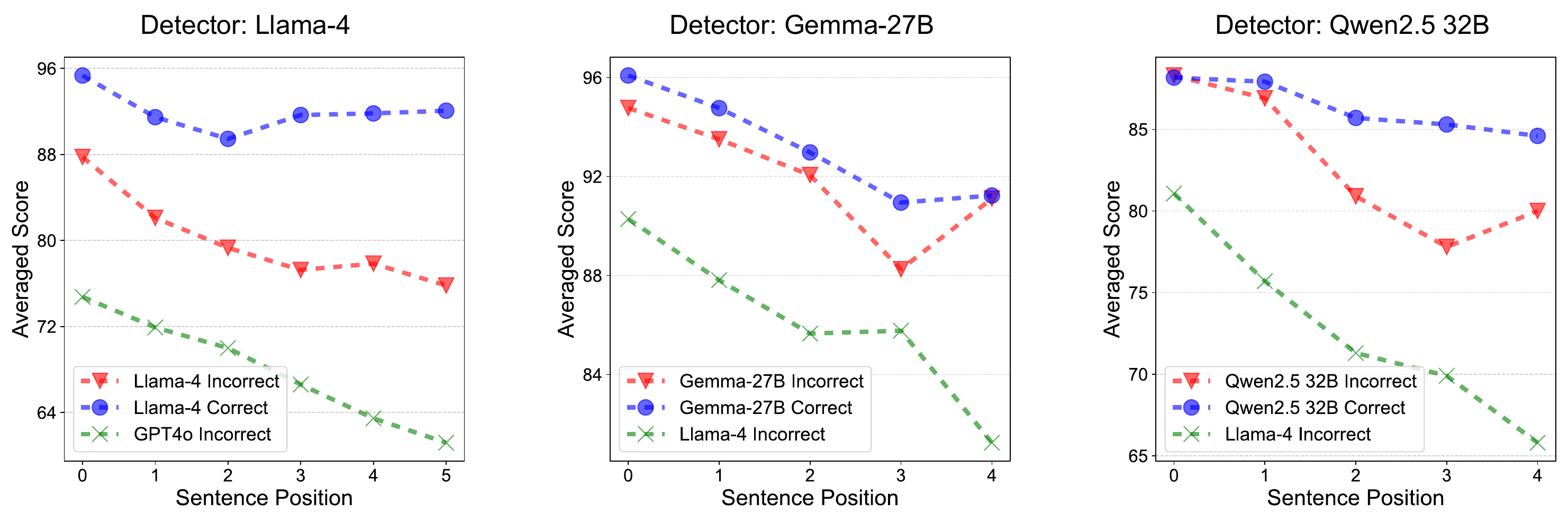}
\caption{\small Detector's output score for their own output captions. }
\label{fig:self_pref_bias_appendix}
\end{figure*}

%% file: figures/score_distribution.tex
\begin{figure*}[h!]
\centering
\includegraphics[width=\linewidth]{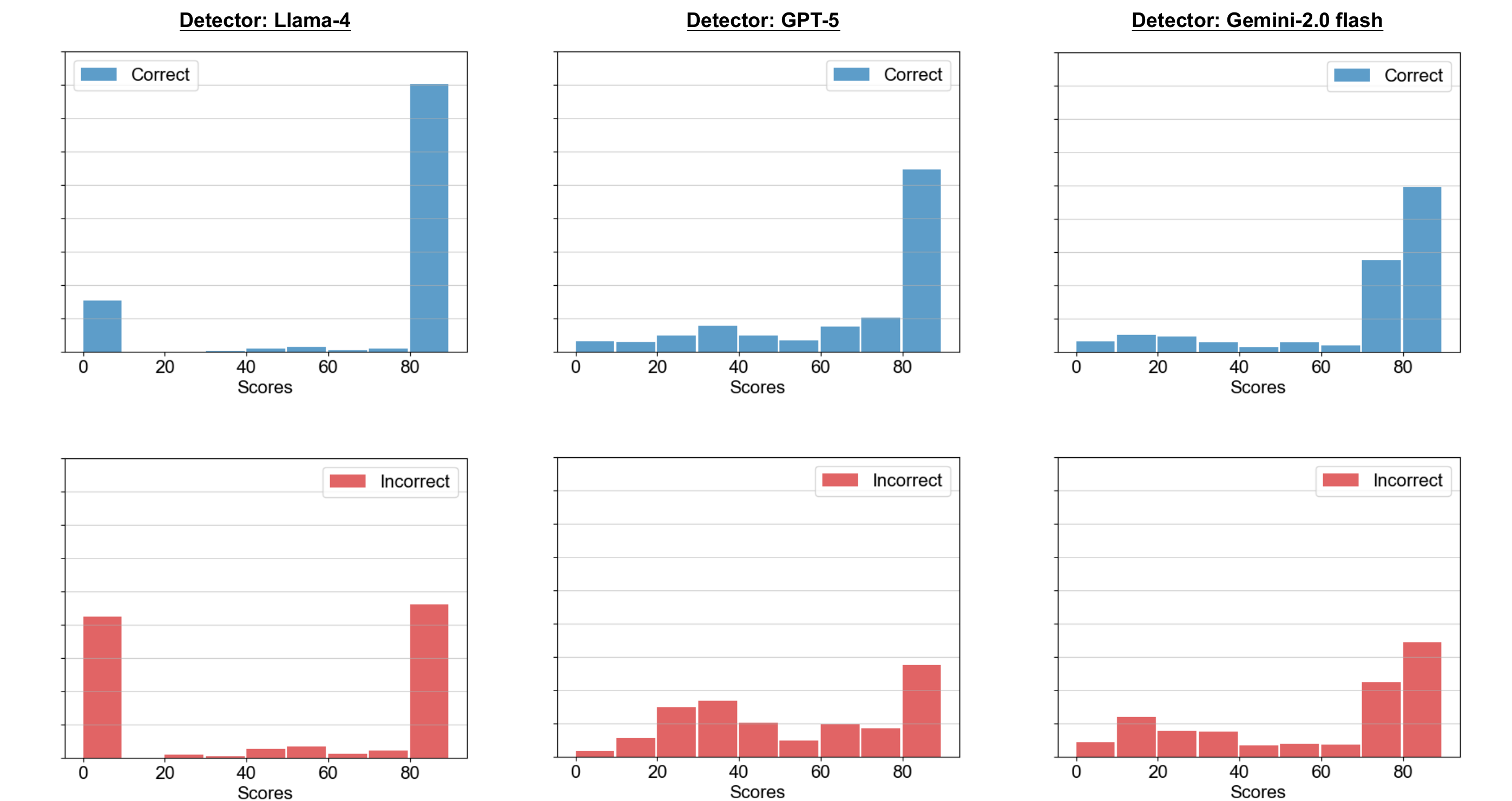}
\caption{\small \textbf{Distributions of evaluators' output scores.} We visualize the evaluators' scores for GPT-4o captions.}
\label{fig:score_dist}
\end{figure*}

%% file: figures/ex_scoring_appendix.tex
\begin{figure*}[t]
\centering
\includegraphics[width=\linewidth]{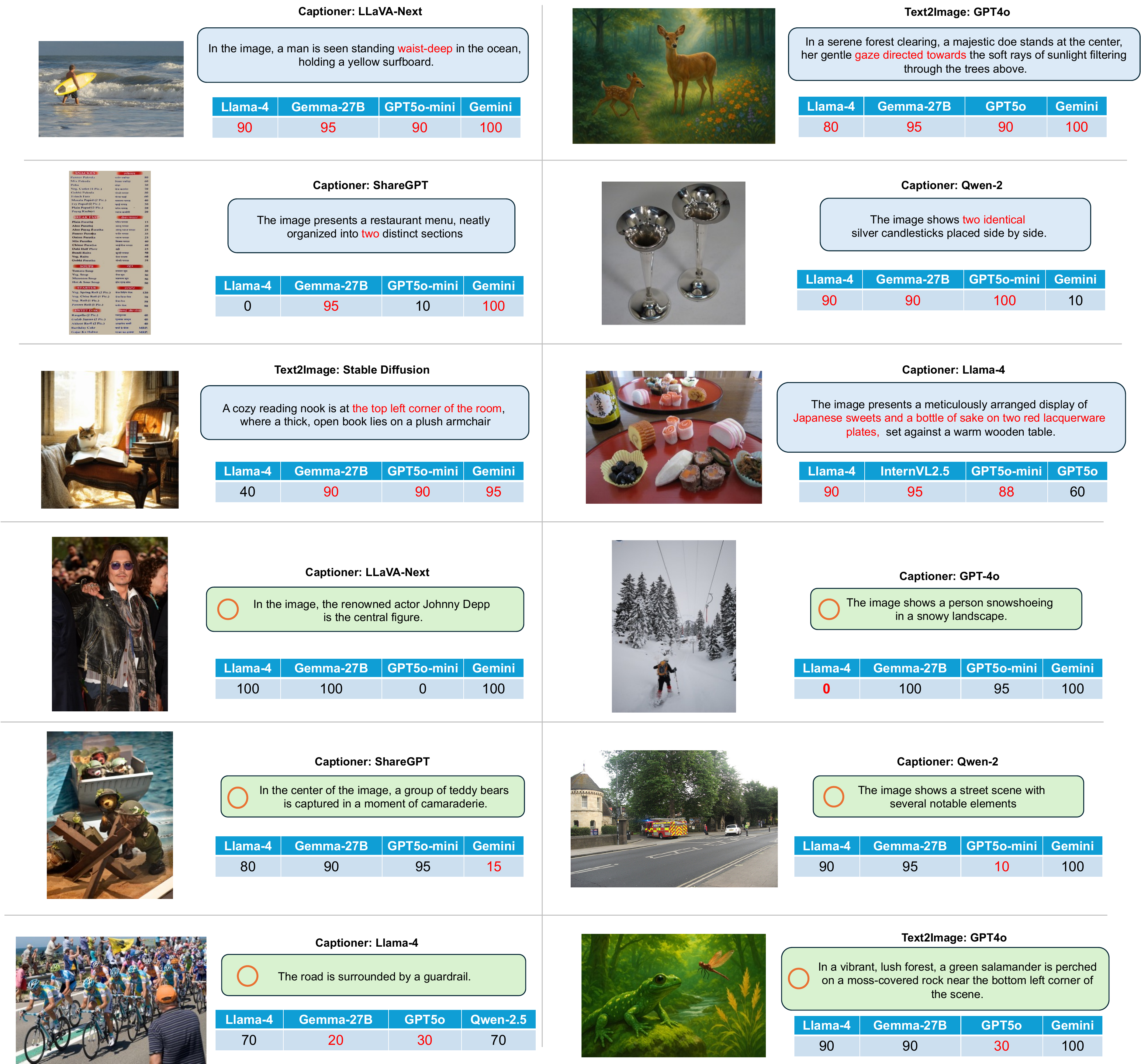}
\caption{\small Examples of input image and sentences with detectors’ correctness scores. Higher scores indicate greater confidence in correctness. We highlight detectors’ errors in red within the text.}
\label{fig:ex_scoring_appendix}
\end{figure*}

%% file: figures/ann_ex_sharegpt.tex
\begin{figure*}[t]
\centering
\includegraphics[width=0.8\linewidth]{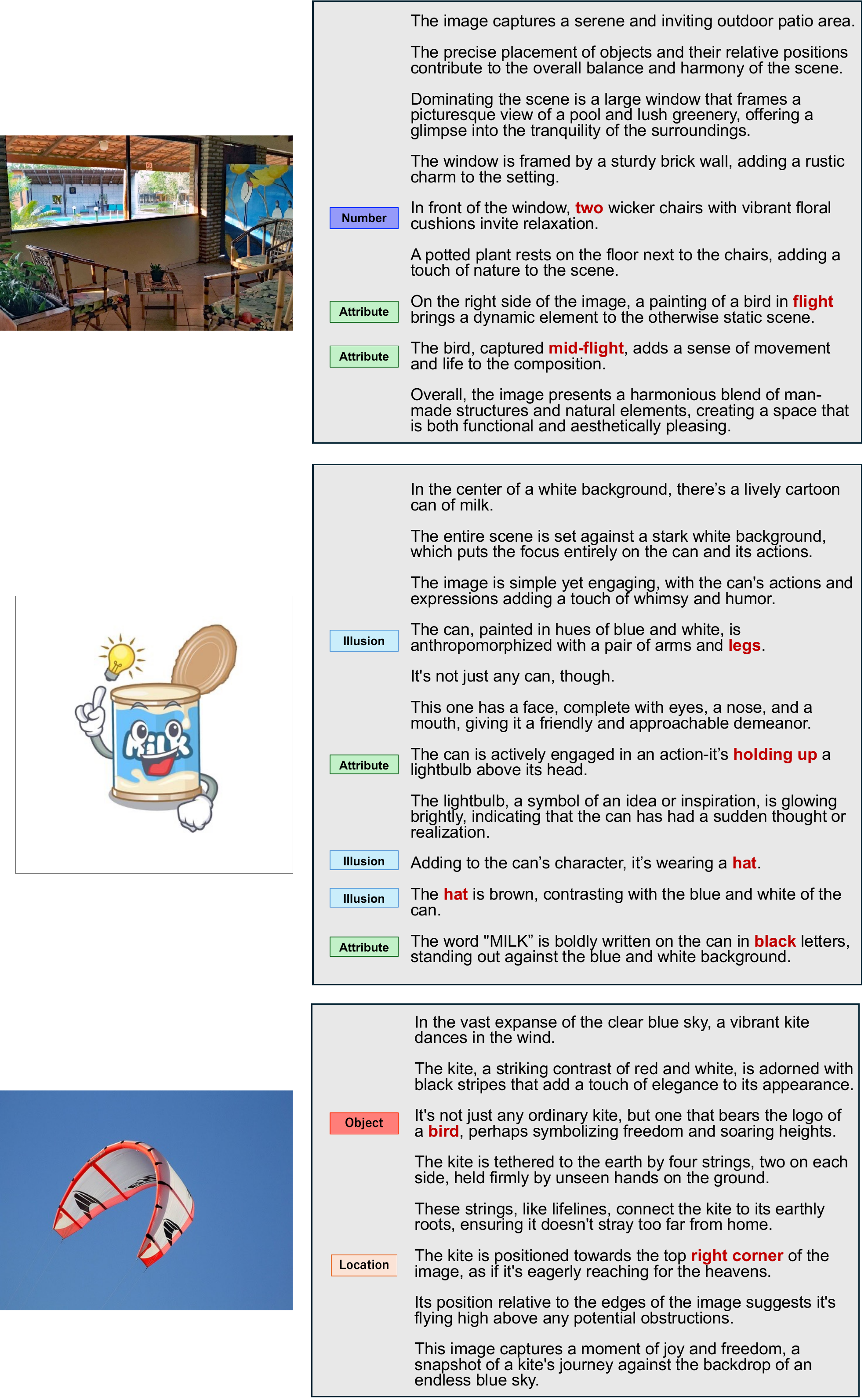}
\caption{Example annotations of Share-GPT.}
\label{fig:error_type_ex_sharegpt}
\end{figure*}

%% file: figures/ann_ex_llava.tex
\begin{figure*}[t]
\centering
\includegraphics[width=0.85\linewidth]{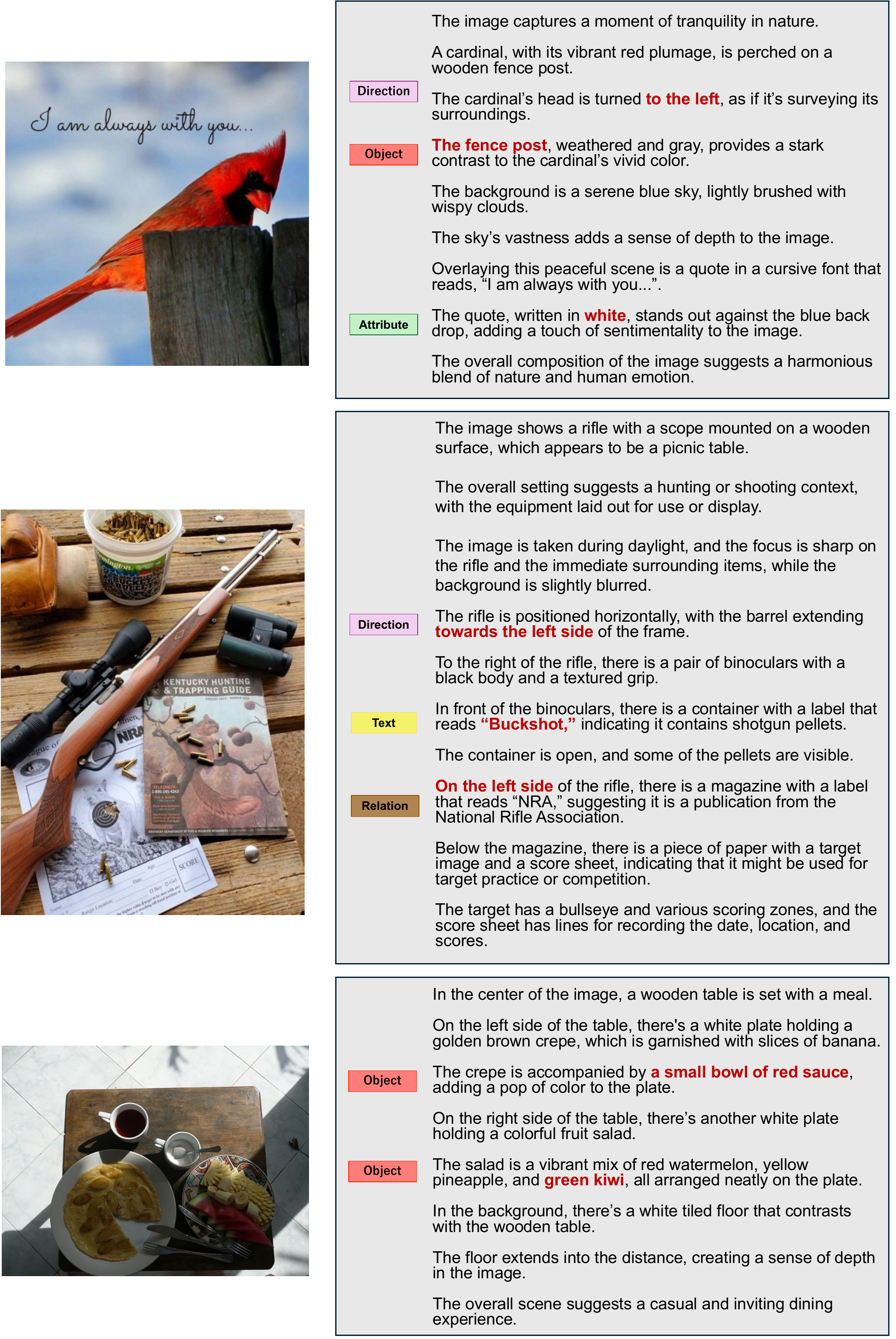}
\caption{Example annotations of LLaVA.}
\label{fig:error_type_ex_llava}
\end{figure*}

%% file: figures/ann_ex_qwen.tex
\begin{figure*}[t]
\centering
\includegraphics[width=\linewidth]{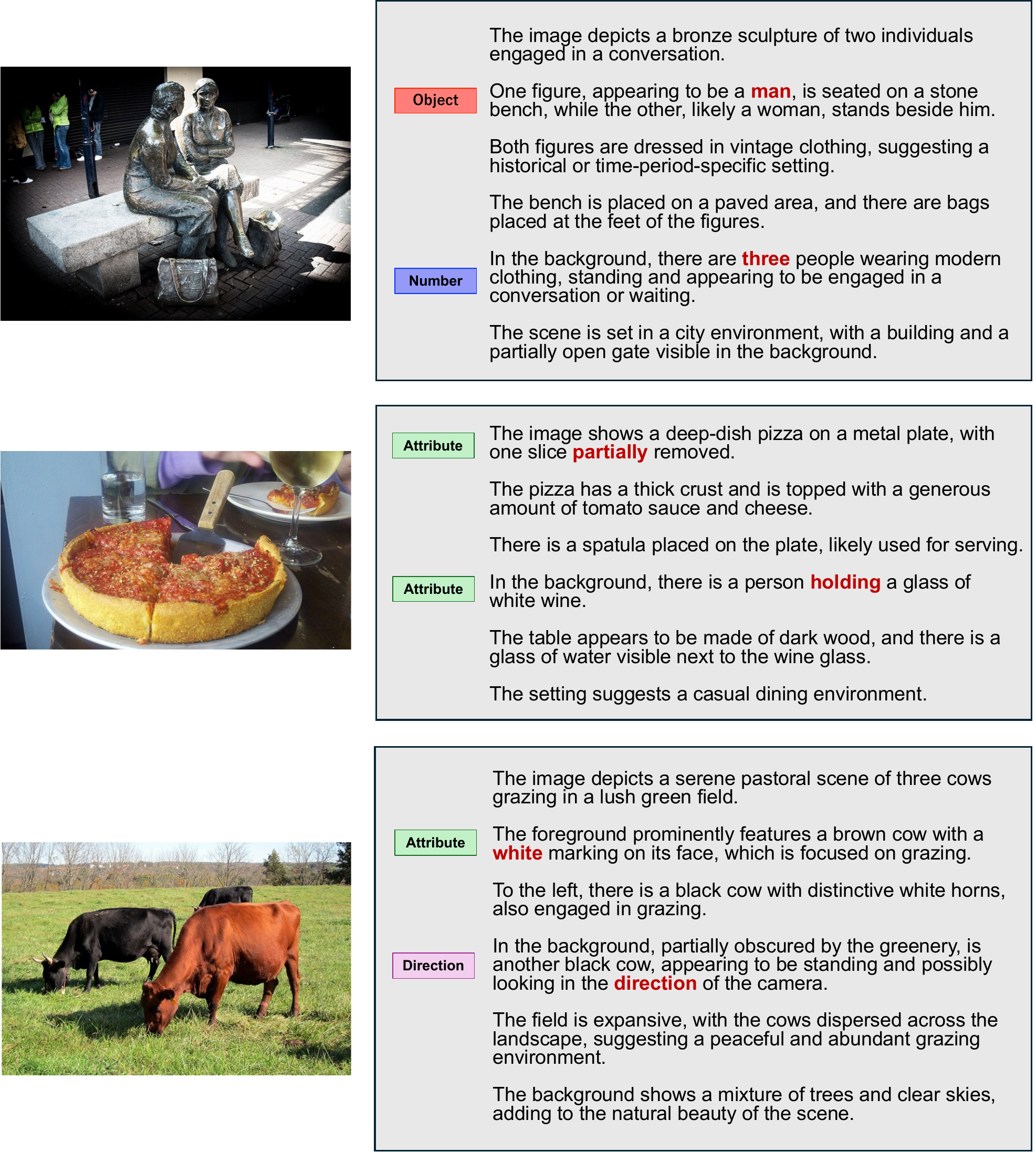}
\caption{Example annotations of Qwen-2.}
\label{fig:error_type_ex_qwen}
\end{figure*}

%% file: figures/ann_ex_gpt4o.tex
\begin{figure*}[t]
\centering
\includegraphics[width=\linewidth]{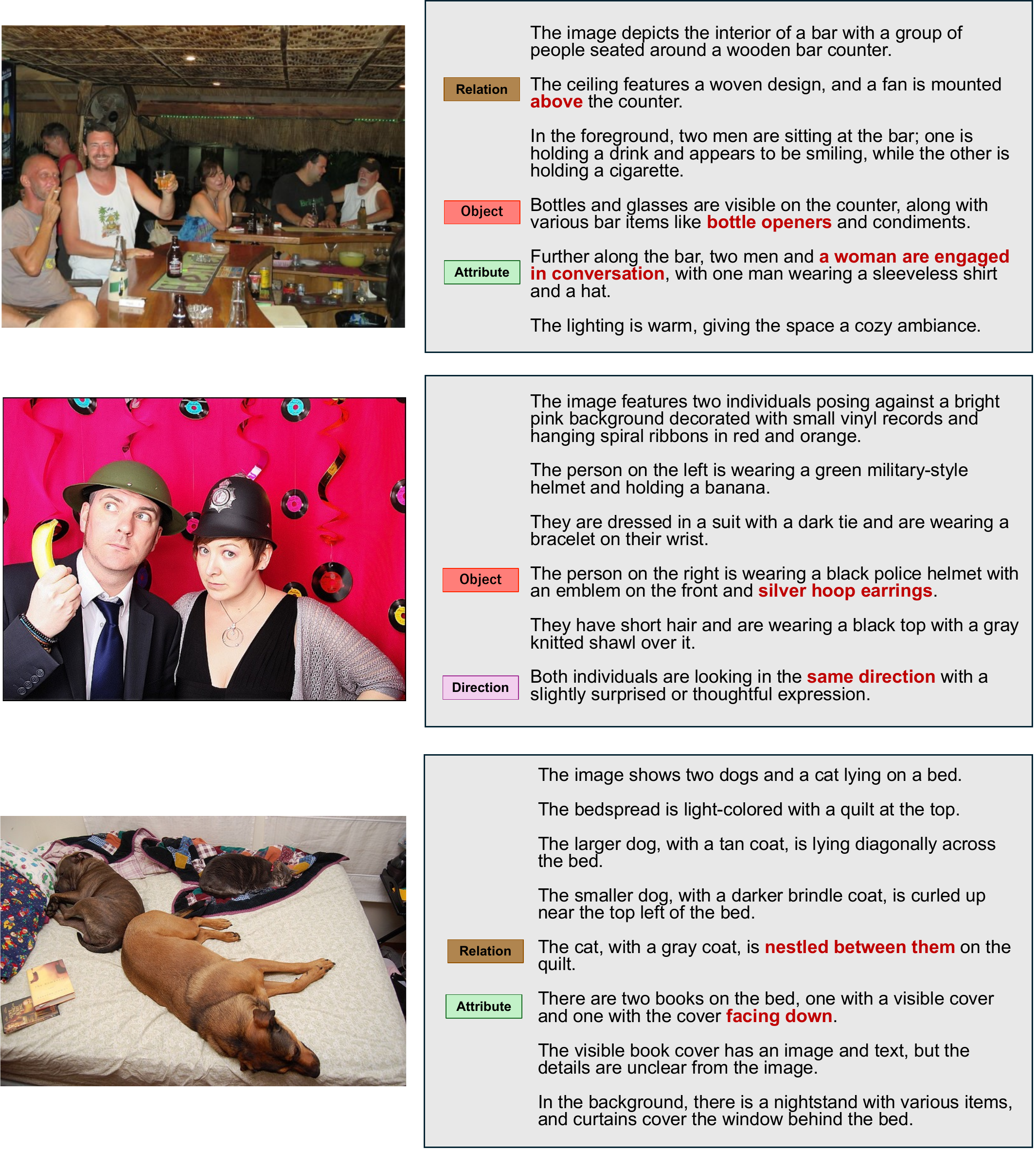}
\caption{Example annotations of GPT4o.}
\label{fig:error_type_ex_gpt4o}
\end{figure*}

%% file: figures/ann_ex_cogvlm.tex
\begin{figure*}[t]
\centering
\includegraphics[width=\linewidth]{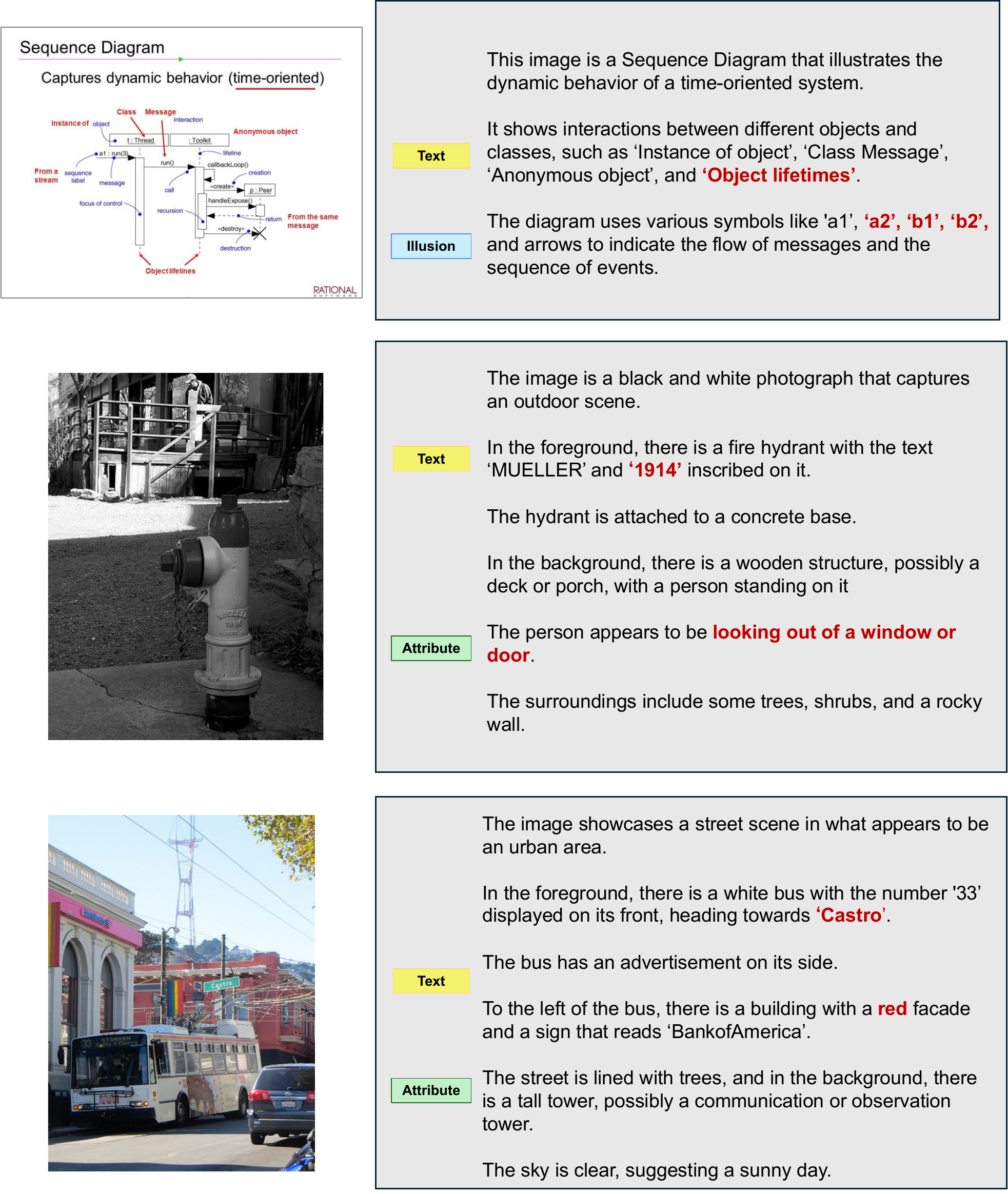}
\caption{Example annotations of CogVLM.}
\label{fig:error_type_ex_cogvlm}
\end{figure*}

%% file: figures/ann_ex_llama-4.tex
\begin{figure*}[t]
\centering
\includegraphics[width=0.85\linewidth]{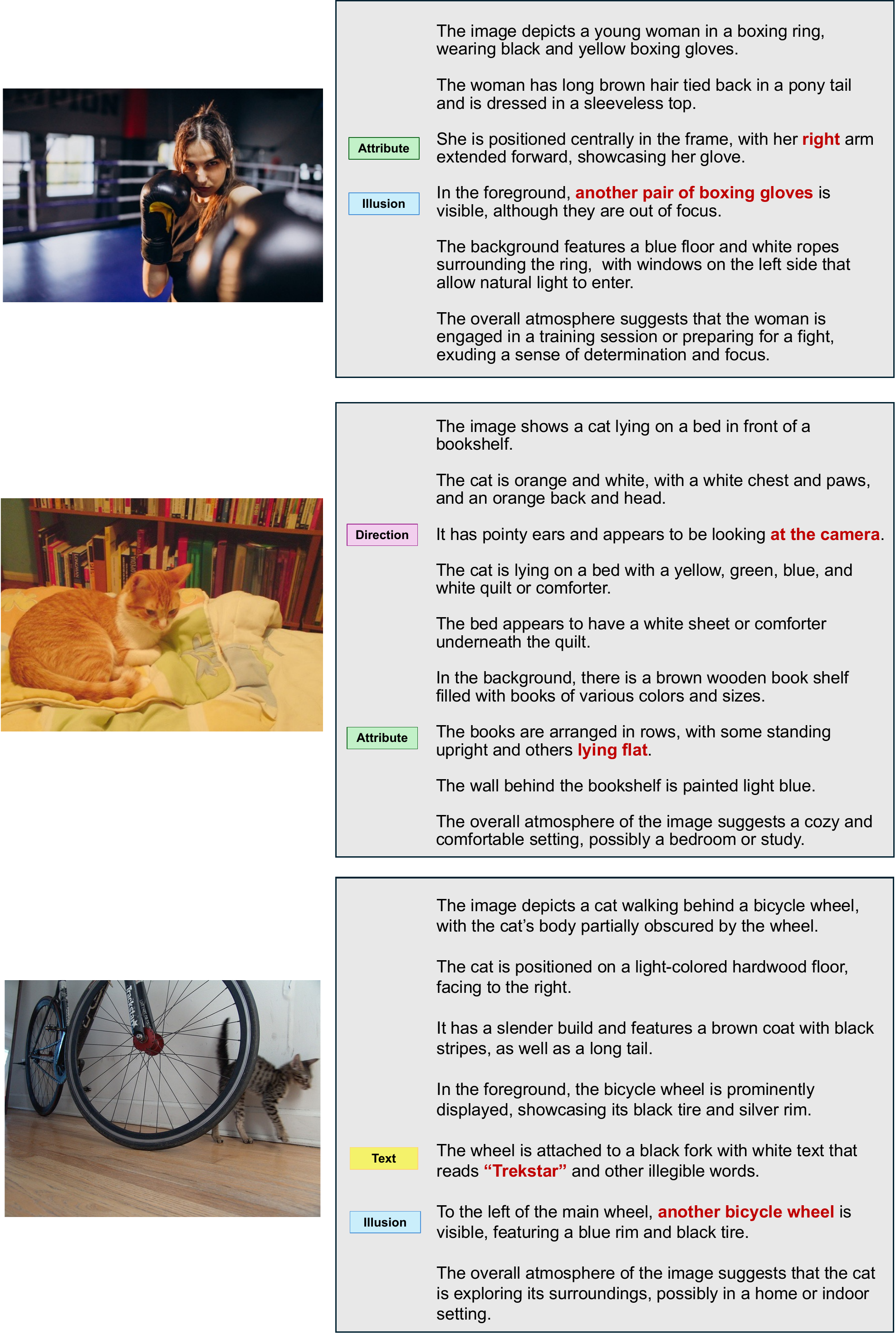}
\caption{Example annotations of Llama-4.}
\label{fig:error_type_ex_llama4}
\end{figure*}

%% file: figures/ann_ex_sd.tex
\begin{figure*}[t]
\centering
\includegraphics[width=\linewidth]{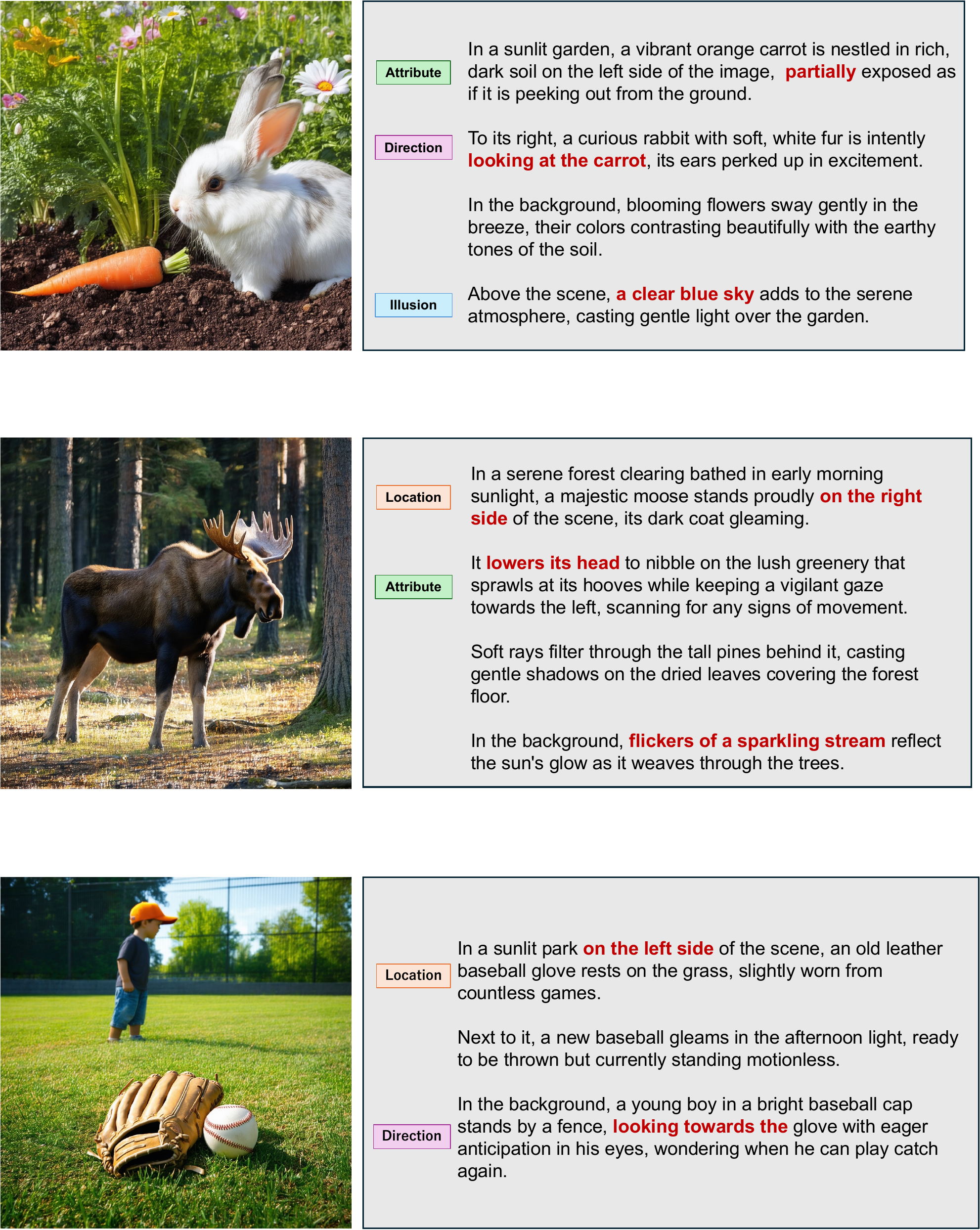}
\caption{Example annotations of Stable Diffusion.}
\label{fig:error_type_ex_sd}
\end{figure*}

%% file: figures/ann_ex_gpt_gen.tex
\begin{figure*}[t]
\centering
\includegraphics[width=\linewidth]{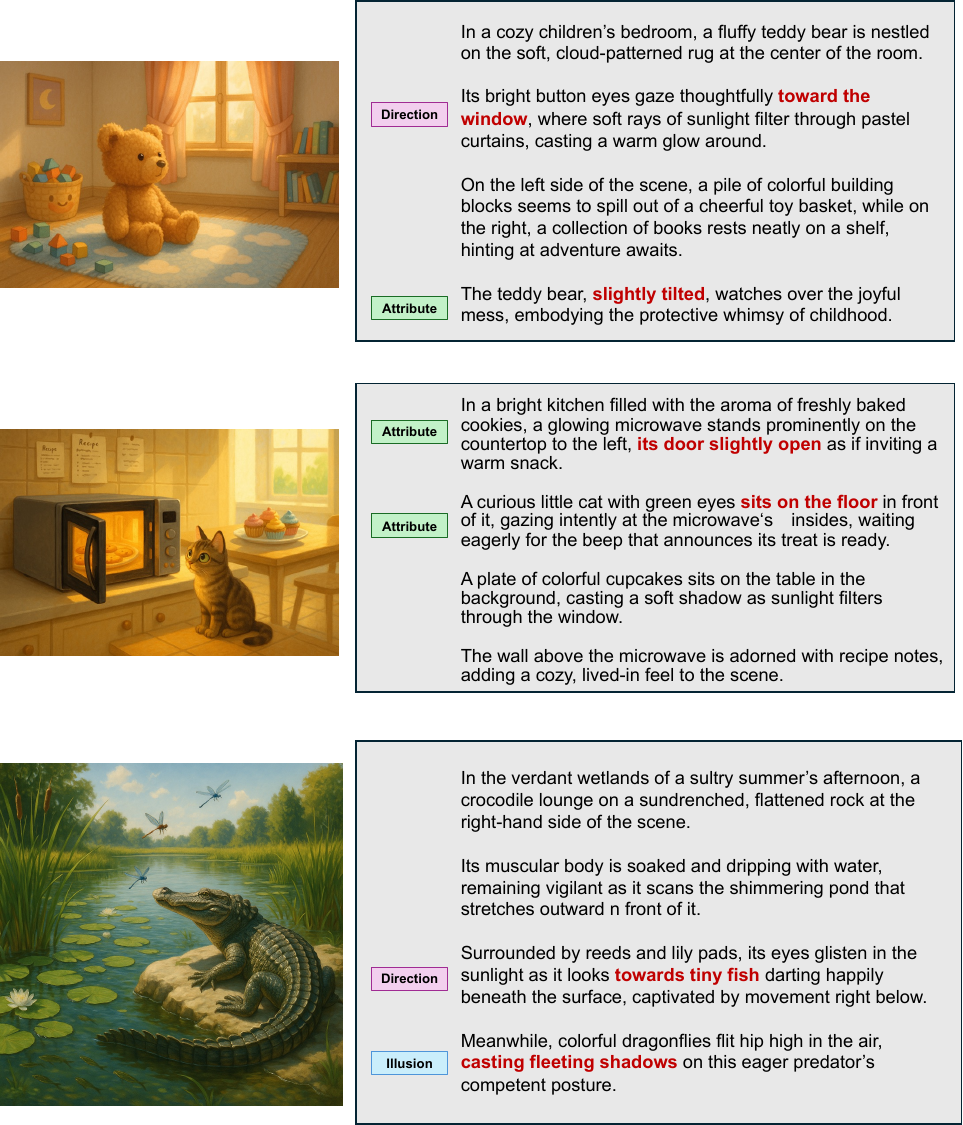}
\caption{Example annotations of GPT-Gen.}
\label{fig:error_type_ex_gptgen}
\end{figure*}